\documentclass{article}

\PassOptionsToPackage{numbers,sort&compress}{natbib}
\usepackage[main, preprint]{neurips_2026}

\usepackage[utf8]{inputenc}
\usepackage[T1]{fontenc}
\usepackage{microtype}
\usepackage{graphicx}
\usepackage{booktabs}
\usepackage{hyperref}
\usepackage{url}
\usepackage{amsmath}
\usepackage{mathtools}
\usepackage{amssymb}
\usepackage{amsfonts}
\usepackage{amsthm}
\usepackage{nicefrac}
\usepackage{needspace}
\usepackage{xcolor}
\usepackage{multirow}
\usepackage{enumitem}
\usepackage{tikz}
\usepackage{tcolorbox}
\usepackage[ruled,vlined,linesnumbered,noend]{algorithm2e}

\hypersetup{
  colorlinks=true,
  linkcolor=blue!55!black,
  citecolor=blue!55!black,
  urlcolor=red!60!black
}

\makeatletter
\renewcommand{\@noticestring}{%
  Preprint. Correspondence to Yao Shu \texttt{<yaoshu@hkust-gz.edu.cn>}.%
}
\makeatother

\graphicspath{{workspace/figs/}}

\usepackage{amsmath,amsfonts,bm}

\def\1{\bm{1}}

\def\rvepsilon{{\mathbf{\epsilon}}}

\def\rvu{{\mathbf{i}}}

\def\rvu{{\mathbf{u}}}

\def\rvx{{\mathbf{x}}}

\def\rvz{{\mathbf{z}}}

\def\rmA{{\mathbf{A}}}

\def\rmC{{\mathbf{C}}}
\def\rmD{{\mathbf{D}}}
\def\rmE{{\mathbf{E}}}

\def\rmH{{\mathbf{H}}}
\def\rmI{{\mathbf{I}}}

\def\rmK{{\mathbf{K}}}

\def\rmM{{\mathbf{M}}}

\def\rmO{{\mathbf{O}}}
\def\rmP{{\mathbf{P}}}

\def\rmR{{\mathbf{R}}}

\def\rmZ{{\mathbf{Z}}}

\def\rmSigma{{\bm{\Sigma}}}

\def\vzero{{\bm{0}}}
\def\vone{{\bm{1}}}

\def\vtheta{{\bm{\theta}}}

\def\vlambda{{\bm{\lambda}}}

\def\vDelta{{\bm{\Delta}}}

\def\veps{{\bm{\epsilon}}}

\def\vb{{\bm{b}}}

\def\vg{{\bm{g}}}

\def\vm{{\bm{m}}}
\def\vn{{\bm{n}}}

\def\vr{{\bm{r}}}

\DeclareMathAlphabet{\mathsfit}{\encodingdefault}{\sfdefault}{m}{sl}
\SetMathAlphabet{\mathsfit}{bold}{\encodingdefault}{\sfdefault}{bx}{n}

\def\gT{{\mathcal{T}}}

\newcommand{\E}{\mathbb{E}}

\newcommand{\R}{\mathbb{R}}

\DeclareMathOperator{\tr}{tr}

\newcommand{\norm}[1]{\left\| #1 \right\|}

\makeatletter
\newcommand{\qty}{%
  \@ifnextchar[{\qty@square}{%
    \@ifnextchar({\qty@paren}{%
      \@ifnextchar|{\qty@bar}{\qty@parenarg}%
    }%
  }%
}
\def\qty@square[#1]{\left[#1\right]}
\def\qty@paren(#1){\left(#1\right)}
\def\qty@bar|#1|{\left|#1\right|}
\def\qty@parenarg#1{\left(#1\right)}
\makeatother

\newcommand{\lamdir}{\lambda}
\newcommand{\lamavg}{\bar{\lambda}}
\newcommand{\ours}{RISE}
\newcommand{\oursfull}{\textit{Retention-Aware Isotropic Shaping for Exact-Gradient Adaptation}}

\tcbset{
  insightbox/.style={
    colback=black!2!white,
    colframe=black!18!white,
    boxrule=0.35pt,
    left=1mm,
    right=1mm,
    top=0.8mm,
    bottom=0.8mm,
    arc=1pt
  },
  theorybox/.style={
    colback=blue!5!white,
    colframe=blue!70!white,
    boxrule=0.5pt,
    left=0.8mm,
    right=1mm,
    top=1mm,
    bottom=1mm,
    arc=1pt
  }
}

\newtheoremstyle{paperplain}
  {3pt plus 1pt minus 1pt}
  {3pt plus 1pt minus 1pt}
  {\fontfamily{LibertinusSerif-TLF}\selectfont\itshape}
  {}
  {\fontfamily{LibertinusSerif-TLF}\selectfont\bfseries}
  {.}
  {0.5em}
  {}
\theoremstyle{paperplain}
\newtheorem{theorem}{Theorem}[section]
\newtheorem{proposition}[theorem]{Proposition}
\newtheorem{lemma}[theorem]{Lemma}
\newtheorem{corollary}[theorem]{Corollary}
\theoremstyle{definition}

\newtheorem{assumption}[theorem]{Assumption}
\theoremstyle{remark}

\title{Why Zeroth-Order Adaptation May Forget Less: A Randomized Shaping Theory}

\author{%
  Yao Shu \quad
  Jian Mu \quad
  Zhongxiang Dai
}

\begin{document}

\maketitle

\begin{abstract}
Continual learning requires new-task adaptation without damaging previously
acquired capabilities. Recent forward-pass and zeroth-order (ZO) results show
that low-query adaptation may retain better than first-order (FO) descent, but
the usual view of ZO as noisy FO estimation does not explain why. We give a
local randomized gradient-shaping analysis: finite differences expose a raw
shape that is mean-aligned with FO, while the norm-matched comparator fixes the
expected squared adaptation norm. Under this controlled comparison, forgetting is
controlled by how the adaptation shape exposes retention curvature. For
norm-matched ZO,
the expected shaped retention curvature obeys an exact identity that preserves the
isotropic retention floor while contracting only the anisotropic component.
Projecting this identity onto the incoming gradient yields the observable FO--ZO
quadratic forgetting gap: ZO improves mean forgetting precisely when the FO
direction has above-average retention curvature, by a query-dependent fraction
of that curvature excess. A practical finite-query accounting separates the
mean mechanism from one-batch sampling and smoothing perturbations. This
mechanism also yields an algorithmic transfer: \ours{} (\oursfull{}) applies the
calibrated ZO shape to exact FO gradients inside parameter blocks. The resulting
stability--plasticity tradeoff is explicit: randomized shaping may reduce the
retention exposure paid by FO, exact gradients remove finite-smoothing bias from
finite-difference ZO, and blockwise sampling supplies many local shaping
directions after one gradient computation.
The blockwise analysis separates mean-step damage from centered random exposure,
showing how block-diagonal curvature, cross-block coupling, and local shaping
scores indicate where this exact-gradient transfer is most likely to be visible.
\end{abstract}

\section{Introduction}
\label{sec:intro}

Continual learning asks a model to absorb new tasks without erasing the
capabilities learned from previous ones. This tension is acute when large models
are repeatedly adapted across domains, users, and deployment conditions: even an
accurate new-task gradient can move parameters through locally fragile
historical-loss directions. Classical methods address this risk with explicit
retention channels: distillation in Learning without Forgetting
\citep{li2016learning}, Fisher-weighted regularization in EWC
\citep{kirkpatrick2017overcoming}, exemplar replay in iCaRL
\citep{rebuffi2017icarl}, and gradient constraints in GEM and A-GEM
\citep{lopezpaz2017gradient,chaudhry2019efficient}. Recent forward-pass and
zeroth-order (ZO) continual-learning work changes the baseline: ZeroFlow
\citep{feng2025zeroflow} and forward-only adaptation
\citep{chen2025forwardonly} show that forward passes can mitigate forgetting,
while ZO-FC reports a ZO/FO stability--plasticity tradeoff in adapter/head
training \citep{yu2025more}. MeZO establishes forward-pass language-model
fine-tuning \citep{malladi2023fine}, and the ZO-LLM benchmark broadens the
large-model evidence for memory-efficient ZO adaptation
\citep{zhang2024revisiting}.
These results make the empirical phenomenon visible, but they do not identify
the scale-controlled local mechanism. This paper studies that mechanism after
the incoming signal and adaptation scale are matched; Appendix~\ref{sec:related}
gives the detailed positioning.

Viewing ZO only as noisy FO estimation misses the object seen by forgetting.
Estimation accuracy concerns the incoming gradient; forgetting is governed by
the induced displacement and the retention curvature it exposes. We therefore
compare FO and ZO through \emph{randomized gradient shaping}: a random operator
acts on the incoming gradient before local adaptation. Sec.~\ref{sec:zo-calibration}
separates the practical finite-difference estimator into raw mean alignment,
smoothing error, norm matching, and the mean shrinkage induced by matching
adaptation scale. With these confounders separated, Sec.~\ref{sec:theory} shows
that calibrated ZO preserves the isotropic part of retention curvature while
contracting the anisotropic part. Projecting this contraction onto the incoming
gradient gives a sign-sensitive gap: ZO reduces mean quadratic forgetting
exactly when the FO direction crosses above-average retention curvature, and it
can hurt in the opposite regime. The theory therefore describes a
query-controlled move toward isotropic exposure, not a uniform ZO-dominates-FO
result.

\ours{} (\oursfull{}) transfers this mechanism to exact-gradient adaptation, as
developed in Sec.~\ref{sec:method}. It computes the FO gradient by
backpropagation, removing finite-difference smoothing bias; applies the
calibrated norm-matched ZO shape, importing the retention-curvature mixing that
may lower FO forgetting in favorable curvature-gap regimes; and applies the
shape inside parameter blocks, turning a small number of random directions per
block into many local shaping samples after one gradient computation. These
pieces define a stability--plasticity tradeoff: \ours{} keeps the exact-gradient
plasticity signal missing from raw finite-difference ZO, while norm matching
still shrinks the mean direction relative to plain FO. The same controls shape
the experiments in Sec.~\ref{sec:experiments}: retention gains from
exact-gradient shape transfer must be separated from smaller steps, mean
scaling, generic covariance noise, and matched-budget optimization effects.

This paper makes four contributions. First, it formulates local forgetting as
shaped adaptation: after the incoming direction is fixed, FO and ZO can be
compared by the quadratic forgetting damage of their induced displacements under
matched expected squared adaptation norm. Second, it calibrates finite-difference
ZO for this comparison by separating raw mean alignment, finite-smoothing
residuals, raw norm inflation, norm-matched scaling, and the resulting mean
shrinkage. Third, it proves the main mechanism: norm-matched ZO preserves the
isotropic retention floor, contracts anisotropic retention curvature, and yields
an exact FO--ZO quadratic-forgetting gap governed by the directional
retention-curvature excess. Fourth, it transfers the calibrated shape to exact
gradients through \ours{}, and evaluates the resulting stability--plasticity
tradeoff with controlled mechanism tests and end-to-end continual-learning
benchmarks. The first step is therefore to make precise what historical loss
sees during a local adaptation.

\section{Local Forgetting as Shaped Adaptation}
\label{sec:setup}

The mechanism above needs a comparison object that is independent of how an
incoming direction was estimated. Historical loss observes the displacement
caused by adaptation, not whether that displacement came from backpropagation or
finite differences. We therefore fix one adaptation stage and compare FO and ZO
at the level where forgetting is measured: the local adaptation displacement.
This local view removes trajectory confounders such as data order, optimizer
state, higher-order drift, and learning-rate schedules. The displacement may be
one optimizer step or a short net adaptation, as long as it stays in the Taylor
neighborhood of the historical loss. At that level, forgetting reduces to a
quadratic damage functional, and FO--ZO comparison becomes a question about the
retention curvature exposed by each induced displacement.

\subsection{From Historical Loss to Quadratic Damage}
\label{sec:setup-forgetting}

Fix an adaptation stage and suppress its index. Let $\vtheta\in\R^d$ with
$d\geq2$ be the parameter before adapting to the incoming loss $f$, and let
$\mathcal{L}$ be the historical loss over past data or tasks. For a local
adaptation displacement $\vDelta\in\R^d$, write
$\vtheta^{+}=\vtheta+\vDelta$ and define induced forgetting as
\begin{equation}
  \mathfrak{F}(\vDelta)
  \triangleq
  \mathcal{L}(\vtheta^{+})-\mathcal{L}(\vtheta)
  =
  \mathcal{L}(\vtheta+\vDelta)-\mathcal{L}(\vtheta).
  \label{eq:forgetting-functional}
\end{equation}
This definition does not specify the optimizer. It keeps only what the
historical loss observes: the displacement and the retention curvature exposed
by that displacement. In the
approximately consolidated regime, assume
$\norm{\nabla\mathcal{L}(\vtheta)}\leq\varepsilon$ for small
$\varepsilon\geq0$. When the quadratic object is the true Taylor term, write
$\rmH_{\mathcal{L}}\triangleq\nabla^2\mathcal{L}(\vtheta)$ and assume the
historical-loss Hessian is locally Lipschitz in the Taylor neighborhood. Then
\begin{equation}
\begin{aligned}
  \mathfrak{F}(\vDelta)
  &=
  \langle\nabla\mathcal{L}(\vtheta),\vDelta\rangle
  +\frac{1}{2}\vDelta^\top\rmH_{\mathcal{L}}\vDelta
  +\mathcal{O}(\norm{\vDelta}^3) \\
  &=
  \frac{1}{2}\vDelta^\top\rmH_{\mathcal{L}}\vDelta
  +\mathcal{O}(\varepsilon\norm{\vDelta}+\norm{\vDelta}^3).
\end{aligned}
\label{eq:forgetting-expansion}
\end{equation}
Thus, up to lower-order terms, the true Taylor contribution is governed by the
historical-loss Hessian along the adaptation direction. The main theory writes
$\rmH$ for the retention curvature used in the quadratic analysis: it may be
$\rmH_{\mathcal{L}}$ in the true-Hessian reading, or a fixed symmetric
retention proxy such as a Fisher, Gauss--Newton, or empirical sensitivity matrix.
In the proxy reading, the next display is a surrogate damage functional rather
than the Taylor equality in \eqref{eq:forgetting-expansion}. The resulting
comparison object is
\begin{equation}
  \mathcal{Q}(\vDelta)
  \triangleq
  \frac{1}{2}\vDelta^\top \rmH\vDelta .
  \label{eq:quadratic-forgetting}
\end{equation}
Introducing $\mathcal{Q}$ does not turn continual learning into a
retention-only objective. Standard methods also separate stability from
plasticity by protecting past knowledge with regularizers such as EWC
\citep{kirkpatrick2017overcoming}, distillation as in Learning without
Forgetting \citep{li2016learning}, replay as in iCaRL
\citep{rebuffi2017icarl}, or gradient constraints as in GEM and A-GEM
\citep{lopezpaz2017gradient,chaudhry2019efficient}, while another term drives
the incoming task.
The present analysis is narrower: once the incoming-loss direction is fixed, the
local question is whether the resulting adaptation shape lowers retention cost.
This scope avoids the trivial zero-displacement solution of minimizing
$\mathcal{Q}(\vDelta)$ over arbitrary displacements. Plasticity enters as a
control: the setup fixes the incoming direction, and
Section~\ref{sec:zo-calibration} separates the raw ZO first-order signal from
the norm-matched squared adaptation scale.

\subsection{Adaptation Shape as the Comparison Object}
\label{sec:setup-shape}

Once forgetting is reduced to $\mathcal{Q}(\vDelta)$, the object to compare is
the displacement placed inside this quadratic form, not the procedure that
produced the direction. Let
$\vg\triangleq\nabla f(\vtheta)\neq\vzero$ be the incoming-loss gradient. A
shape operator $\rmP$ transforms this gradient before it becomes a local
adaptation. With learning rate $\eta>0$,
\begin{equation}
  \vDelta=-\eta\rmP\vg,
  \qquad
  \mathcal{Q}_{\rmP}\triangleq \mathcal{Q}(\vDelta)
  =
  \frac{\eta^2}{2}\vg^\top\rmP^\top\rmH\rmP\vg .
  \label{eq:shaped-damage}
\end{equation}
Here $\mathcal{Q}_{\rmP}$ is the forgetting damage induced by $\rmP$, and
$\rmP^\top\rmH\rmP$ is the retention curvature seen by the shaped incoming
gradient. We call this operator-level transformation \emph{shaping}, and call it
\emph{randomized shaping} when $\rmP$ is random. At this stage $\rmP=\rmI$
gives the FO reference; ZO enters in Section~\ref{sec:zo-calibration} as a
calibrated randomized shaping distribution.

The local comparison also removes scale as an explanation. Because
$\mathcal{Q}$ is quadratic, a method could reduce forgetting merely by producing
a smaller displacement. We therefore match expected squared adaptation norm, but
do not solve an unrestricted minimization over arbitrary curvature-aware
conditioners. If $\rmP$ could be chosen with full knowledge of
$\rmH$, the best rule would steer $\rmP\vg$ toward low-curvature eigenspaces,
which is not the ZO family calibrated in Sec.~\ref{sec:zo-calibration}.
Instead, for a specified curvature-agnostic family $\mathcal{C}$ of shaping
rules, the theory compares
\begin{equation}
  \left\{
  \E\left[\mathcal{Q}_{\rmP}\right]:
  \rmP\in\mathcal{C},\
  \E\left[\norm{\rmP\vg}^{2}\right]=\norm{\vg}^{2}
  \right\}.
  \label{eq:retention-only-problem}
\end{equation}
For this paper, $\mathcal{C}$ contains the FO reference and the ZO shaping
rules calibrated in Section~\ref{sec:zo-calibration}. The expectation is over
shaping randomness and is vacuous for deterministic FO. The norm constraint
prevents smaller expected squared adaptation from being mistaken for a retention
mechanism. ZO therefore needs a rule with two properties: a raw form that aligns
the incoming-loss signal and a normalized form that fits the norm-matched
comparison. Sec.~\ref{sec:zo-calibration} provides that calibration.

\begin{assumption}[Local comparison]
\label{ass:local}
The local comparison in \eqref{eq:retention-only-problem} relies on three
conditions: (i) $\vg\neq\vzero$;
(ii) the stage is in the consolidated local regime, so the first-order and
higher-order terms in \eqref{eq:forgetting-expansion} are lower-order at the
local displacement scale considered; and (iii) $\rmH$ is a fixed symmetric
retention curvature or proxy used in $\mathcal{Q}$. The Gaussian moment
identities in Section~\ref{sec:theory} require only this symmetry. Whenever
$\mathcal{Q}$ is interpreted as nonnegative forgetting damage, and in the
curvature-agnostic benchmark and finite-query deviation bound, we additionally
assume $\rmH\succeq0$ and $\tr(\rmH)>0$.
\end{assumption}

This assumption turns forgetting into a displacement-level comparison, but
\eqref{eq:retention-only-problem} still needs a ZO rule that supplies both the
incoming signal and the matched scale. Finite differences provide that rule:
their raw shape carries the leading first-order signal, and their normalized
shape fits the norm-matched comparison.

\section{Calibrating the ZO Shape}
\label{sec:zo-calibration}

The shaped-adaptation comparison in \eqref{eq:retention-only-problem} uses a
common local form
$-\eta\rmP\vg$, but practical ZO arrives through finite differences rather than
an explicit shape operator. The two-point estimator admits the needed
separation: a raw ZO shape carrying the leading first-order signal, a
finite-smoothing residual that remains a controlled perturbation, and a
norm-matched shape whose expected squared adaptation norm is comparable with FO.

\subsection{From Finite Differences to Raw ZO Shape}
\label{sec:finite-smoothing}

The random directions used by the two-point ZO estimator define the raw shape.
For query count $q$, draw
$\rvz_1,\ldots,\rvz_q\overset{\text{i.i.d.}}{\sim}\mathcal{N}(\vzero,\rmI)$
independently of $\vg$ and $\rmH$; all expectations below are over these
directions. With smoothing radius $\mu>0$, the practical estimator and its
zero-smoothing raw shape are
\begin{equation}
  \widehat{\vg}_{\mu}
  \triangleq
  q^{-1}\sum_{r=1}^{q}
  \frac{f(\vtheta+\mu\rvz_r)-f(\vtheta-\mu\rvz_r)}{2\mu}\,\rvz_r,
  \qquad
  \rmZ \triangleq q^{-1}\sum_{r=1}^{q}\rvz_r\rvz_r^\top .
  \label{eq:two-point-estimator}
\end{equation}
The matrix $\rmZ$ is the leading zero-smoothing shape induced by the sampled
directions. Taylor expansion along the queried line segments separates the
practical estimator into this raw shape applied to the incoming gradient plus a
finite-smoothing residual. Because the same estimator will later be inserted
into a quadratic damage functional, the residual is controlled at both first and
second moments: the first moment justifies the estimator-level reduction to
$\rmZ\vg$, and the second moment controls its contribution to forgetting
damage.

\begin{lemma}[Finite-Smoothing Bridge]
\label{lem:finite-smoothing}
Let $f$ be $L$-smooth on the queried line segments. Then there exists a
remainder vector
$\vr_{\mu}$ such that
\begin{equation}
  \widehat{\vg}_{\mu}
  =
  \rmZ\vg + \vr_{\mu},
  \qquad
  \E\left[\norm{\vr_{\mu}}\right]
  =
  \mathcal{O}(L\mu d^{3/2}),
  \qquad
  \E\left[\norm{\vr_{\mu}}^2\right]
  =
  \mathcal{O}(L^2\mu^2 d^3).
  \label{eq:finite-mu-bridge}
\end{equation}
\end{lemma}

Lemma~\ref{lem:finite-smoothing} identifies $\rmZ\vg$ as the leading shaped
incoming direction and keeps finite smoothing outside the zero-smoothing
curvature mechanism. The raw shape still requires two calibrations: preservation
of the incoming signal in expectation and a squared adaptation norm comparable
with FO.

\subsection{From Raw Alignment to a Norm-Matched Comparator}
\label{sec:mean-norm}

Raw ZO preserves the leading mean but inflates the quadratic scale. The fair
forgetting comparator is therefore not the raw adaptation $\rmZ\vg$ itself, but
the normalized shape $\rmP\vg=\kappa^{-1/2}\rmZ\vg$. This normalization matches
the expected squared adaptation norm with FO and makes the induced mean
shrinkage explicit.

\begin{lemma}[Raw Alignment, Norm Inflation, and Norm Matching]
\label{lem:mean-norm}
For the raw ZO shape $\rmZ$ in \eqref{eq:two-point-estimator},
\begin{equation}
  \E\left[\rmZ\right] = \rmI,
  \qquad
  \E\left[\norm{\rmZ\vg}^2\right] = \kappa\norm{\vg}^2,
  \qquad
  \kappa = \frac{q+d+1}{q}.
  \label{eq:mean-norm}
\end{equation}
For the norm-matched shape $\rmP=\kappa^{-1/2}\rmZ$,
\begin{equation}
  \E[\rmP\vg]
  =
  \sqrt{\frac{q}{q+d+1}}\,\vg,
  \qquad
  \E\left[\norm{\rmP\vg}^{2}\right]
  =
  \norm{\vg}^{2}.
  \label{eq:norm-matched-calibration}
\end{equation}
\end{lemma}

Lemma~\ref{lem:mean-norm} separates the two calibration roles. The identity
$\E[\rmZ]=\rmI$ says that raw ZO preserves the FO incoming signal in the
zero-smoothing leading term. The inflation factor
$\kappa=(q+d+1)/q$ comes from the Wishart-type fourth moment
$\E[\rmZ^2]=\kappa\rmI$, so it is not a tunable scale choice. Because raw ZO
inflates $\E[\norm{\rmZ\vg}^{2}]$, it cannot be the fair quadratic-forgetting
comparator. The comparator is instead $\rmP=\kappa^{-1/2}\rmZ$, which matches
the FO squared norm exactly but shrinks the mean to
$\sqrt{q/(q+d+1)}\,\vg$. This asymmetry is unavoidable: for any random shaped
direction $\rvx$,
$\E\left[\norm{\rvx}^{2}\right]=\norm{\E[\rvx]}^{2}
+\E\left[\norm{\rvx-\E[\rvx]}^{2}\right]$, so exact mean alignment and exact
expected squared-norm matching force zero shaping variance. Scaled-FO and
covariance-matched noise controls are therefore needed before attributing a
retention change to curvature shaping.

After this calibration, the curvature theory compares only two shape operators:
\begin{equation}
  \rmP=\rmI\quad\text{(FO)},\qquad
  \rmP=\kappa^{-1/2}\rmZ\quad\text{(ZO)} .
\label{eq:shape-subcases}
\end{equation}
Here $\rmZ$ justifies the raw first-order signal, while
$\rmP=\kappa^{-1/2}\rmZ$ is the norm-matched ZO shape used in the retention
comparison. The finite-smoothing residual remains outside the zero-smoothing
curvature mechanism and reappears as the additive perturbation in
\eqref{eq:practical-finite-mu-accounting} when the mean gap is translated to a
practical finite-difference batch. With raw alignment, norm inflation,
norm-matched mean shrinkage, and expected squared norm separated, the
zero-smoothing retention mechanism reduces to the expected shaped retention
curvature defined in \eqref{eq:mean-shaped-metric}.

\section{Why ZO May Forget Less: Randomized Gradient Shaping}
\label{sec:theory}

The calibration in Sec.~\ref{sec:zo-calibration} leaves FO and norm-matched ZO
on a common local scale:
raw ZO carries the first-order signal, finite smoothing is separated as a
remainder, and norm matching fixes the expected squared adaptation norm while
making mean shrinkage explicit. The remaining zero-smoothing difference is the
retention curvature exposed by the shaped adaptation.

\subsection{Expected Shaped Retention Curvature}
\label{sec:anisotropy}

For a shaped adaptation $-\eta\rmP\vg$, the historical loss sees
$\rmP^\top\rmH\rmP$ rather than $\rmH$. The expectation of this induced
curvature matrix is the central object:
\begin{equation}
  \bar{\rmH}_{q}
  \triangleq
  \E\left[\rmP^\top\rmH\rmP\right],
  \qquad
  \lamavg
  \triangleq
  \frac{\tr(\rmH)}{d}.
  \label{eq:mean-shaped-metric}
\end{equation}
Here $\lamavg$ is the average retention curvature, and $q$ is the number of ZO
directions used by the norm-matched shape in \eqref{eq:shape-subcases}. If the
retention curvature were known and positive definite, a whitening-like reference
$\rmP_\star=\sqrt{\lamavg}\,\rmH^{-1/2}\rmO$ with $\rmO^\top\rmO=\rmI$ would
realize $\rmP_\star^\top\rmH\rmP_\star=\lamavg\rmI$. This full-information
reference only names the isotropic target; the actual ZO shape has no access to
the eigensystem of $\rmH$. Finite-query ZO can now be evaluated by whether it
moves the exposed retention curvature toward this isotropic target without
knowing $\rmH$.

\begin{theorem}[Expected Curvature Identity for Norm-Matched ZO]
\label{thm:anisotropy}
For the norm-matched ZO shape in \eqref{eq:shape-subcases}, let
$\tau=d/(q+d+1)$. For any fixed symmetric $\rmH$,
\begin{equation}
  \bar{\rmH}_{q}
  =
  (1-\tau)\rmH + \tau\lamavg \rmI.
  \label{eq:anisotropy-attenuation}
\end{equation}
\end{theorem}

Theorem~\ref{thm:anisotropy} is the mechanism backbone. It says that
norm-matched ZO does not uniformly shrink retention curvature. Instead, it
leaves the isotropic component $\lamavg\rmI$ unchanged and multiplies only the
traceless anisotropic component $\rmH-\lamavg\rmI$ by $1-\tau$. The standard
normal direction distribution supplies this closed-form moment identity; no
symmetry of the neural network, loss landscape, or optimizer state is assumed.
The query
count enters only through $\tau=d/(q+d+1)$: smaller $q$ gives stronger
mixing toward isotropic curvature, while larger $q$ approaches the FO curvature.

\begin{corollary}[Spectral Equalization]
\label{cor:spectral}
Let $\lambda_1,\ldots,\lambda_d$ be the eigenvalues of $\rmH$, and let
$\lambda'_1,\ldots,\lambda'_d$ be the eigenvalues of
$\E\left[\rmP^\top\rmH\rmP\right]$. For the norm-matched ZO shape in
\eqref{eq:shape-subcases} and any fixed symmetric $\rmH$,
\begin{equation}
  \lambda_i-\lambda'_i
  =
  \tau\left(\lambda_i-\lamavg\right),
  \qquad
  \lambda'_i
  =
  (1-\tau)\lambda_i + \tau\lamavg,
  \qquad i=1,\ldots,d.
  \label{eq:spectral-equalization}
\end{equation}
Consequently, the mean eigenvalue is fixed and every eigen-deviation from
$\lamavg$ is contracted by the factor $1-\tau$.
\end{corollary}

Corollary~\ref{cor:spectral} gives the eigen-level interpretation of the same
mechanism. Every eigenvalue sheds a $\tau$ fraction of its signed deviation from
the mean. Large retention-curvature eigenvalues move down, small ones move up,
and the average is preserved, so the spectrum becomes less anisotropic without
changing its trace.

\subsection{Quantifying the FO--ZO Forgetting Reduction}
\label{sec:direct-damage}

The expected-curvature identity in Theorem~\ref{thm:anisotropy} explains how
norm-matched ZO reshapes retention curvature. Its forgetting effect is the
projection of that shaped curvature onto the fixed incoming gradient. Define the
directional retention curvature
\begin{equation}
  \lamdir
  \triangleq
  \frac{\vg^\top \rmH\vg}{\norm{\vg}^{2}},
  \qquad
  \lamavg
  =
  \frac{\tr(\rmH)}{d}.
  \label{eq:curvature-summaries}
\end{equation}
The FO baseline quadratic damage and the norm-matched ZO quadratic damage are
\begin{equation}
  \mathcal{Q}^{\text{FO}}
  \triangleq
  \mathcal{Q}(-\eta\vg)
  =
  \frac{\eta^2}{2}\norm{\vg}^{2}\lamdir,
  \qquad
  \mathcal{Q}_q^{\text{ZO}}
  \triangleq
  \mathcal{Q}(-\eta\rmP\vg)
  =
  \frac{\eta^2}{2}\vg^\top\rmP^\top\rmH\rmP\vg .
  \label{eq:fo-zo-damage-definitions}
\end{equation}
These quantities use the same incoming gradient before shaping and matched
expected squared adaptation norm after shaping. Thus the comparison removes
smaller-step explanations, while the explicit mean shrinkage in
\eqref{eq:norm-matched-calibration} is handled by scaled-FO controls. Any
remaining mean forgetting reduction comes from the shaped retention curvature in
Theorem~\ref{thm:anisotropy}.

\begin{theorem}[FO--ZO Forgetting-Reduction Identity]
\label{thm:direct-damage}
For the norm-matched ZO shape in \eqref{eq:shape-subcases}, let
$\tau=d/(q+d+1)$. For any fixed symmetric $\rmH$,
\begin{equation}
  \begin{aligned}
  \mathcal{Q}^{\text{FO}}
  -
  \E\left[\mathcal{Q}_q^{\text{ZO}}\right]
  =
  \frac{\eta^2}{2}\tau\norm{\vg}^{2}\left(\lamdir-\lamavg\right),
  \qquad
  \E\left[\mathcal{Q}_q^{\text{ZO}}\right]
  =
  \frac{\eta^2}{2}\norm{\vg}^{2}
  \left((1-\tau)\lamdir+\tau\lamavg\right).
  \end{aligned}
  \label{eq:direct-damage-gap}
\end{equation}
If $\lamdir>0$, the corresponding relative reduction is
\begin{equation}
  \frac{
  \mathcal{Q}^{\text{FO}}
  -
  \E\left[\mathcal{Q}_q^{\text{ZO}}\right]
  }{
  \mathcal{Q}^{\text{FO}}
  }
  =
  \tau\left(1-\frac{\lamavg}{\lamdir}\right).
  \label{eq:relative-forgetting-reduction}
\end{equation}
\end{theorem}

When $\rmH\succeq0$ and $\mathcal{Q}$ is read as forgetting damage,
Theorem~\ref{thm:direct-damage} gives the promised quantitative comparison:
norm-matched ZO reduces mean quadratic forgetting exactly when the current
gradient points through above-average retention curvature, $\lamdir>\lamavg$.
The amount saved is $(\eta^2/2)\tau\norm{\vg}^{2}(\lamdir-\lamavg)$, and the
fraction of FO damage removed is $\tau(1-\lamavg/\lamdir)$ whenever
$\lamdir>0$. Equivalently, $\lamavg\norm{\vg}^{2}$ is the isotropic forgetting
floor, while
$\norm{\vg}^{2}(\lamdir-\lamavg)=\vg^\top(\rmH-\lamavg\rmI)\vg$ is the signed
directional excess above that floor; ZO shaping leaves the floor unchanged and
attenuates only this excess by the factor $1-\tau$. If $\lamdir<\lamavg$, the
same expression is negative and norm-matched ZO increases the mean local damage,
so the mechanism controls high-curvature exposure by giving up the lucky case in
which FO already moves through below-average retention curvature. The result is
therefore a geometry-dependent risk-control statement, not a uniform dominance
theorem.

\paragraph{Practical finite-query ZO.}
Theorem~\ref{thm:direct-damage} is the mean zero-smoothing gap. A practical
finite-difference ZO batch under the same norm-matched calibration adds two
separate perturbations. With
$\widehat{\rvx}_{\mu}\triangleq\kappa^{-1/2}\widehat{\vg}_{\mu}$ and
$\widehat{\mathcal{Q}}_{q,\mu}^{\text{ZO}}
\triangleq\mathcal{Q}(-\eta\widehat{\rvx}_{\mu})$, the accounting is
\begin{align}
  G_q
  &\triangleq
  \mathcal{Q}^{\text{FO}}
  -
  \E\left[\mathcal{Q}_q^{\text{ZO}}\right]
  =
  \frac{\eta^2}{2}\tau\norm{\vg}^{2}\left(\lamdir-\lamavg\right),
  \label{eq:mean-gap-signal}\\
  \mathcal{Q}^{\text{FO}}
  -
  \widehat{\mathcal{Q}}_{q,\mu}^{\text{ZO}}
  &=
  G_q
  -
  \left(
  \mathcal{Q}_q^{\text{ZO}}
  -
  \E\left[\mathcal{Q}_q^{\text{ZO}}\right]
  \right)
  -
  \left(
  \widehat{\mathcal{Q}}_{q,\mu}^{\text{ZO}}
  -
  \mathcal{Q}_q^{\text{ZO}}
  \right).
  \label{eq:practical-finite-mu-accounting}
\end{align}
Thus the practical ZO gap has the same mechanism signal as
Theorem~\ref{thm:direct-damage}. The two parentheses in
\eqref{eq:practical-finite-mu-accounting} are the centered
finite-query deviation and the finite-smoothing perturbation, respectively; the
zero-smoothing case drops the last parenthesis. A realized benefit requires
$G_q$ to dominate both perturbations. Proposition~\ref{prop:finite-query-deviation}
and Corollary~\ref{cor:finite-smoothing-damage} in
Appendix~\ref{app:practical-zo-gap} give the corresponding high-probability
one-batch deviation bound and expected finite-smoothing damage perturbation.

\subsection{When Curvature Is Unknown: ZO Moves Toward the Blind-Optimal Isotropic Exposure}
\label{sec:theory-scope}

The fixed-$\rmH$ results in
Secs.~\ref{sec:anisotropy}--\ref{sec:direct-damage} explain the actual FO--ZO
forgetting gap for a given local retention geometry. A complementary scope
question arises when curvature information is absent. Scalar norm matching fixes
only the total squared norm; the full exposure matrix matters once unknown
curvature can choose directions. The benchmark below is retention-only and acts
on second moments, not a joint plasticity--retention optimum under exact mean
constraints. Let $\rvx=\rmP\vg$, $\vDelta=-\eta\rvx$, and
$\rmM=\E\left[\rvx\rvx^\top\right]$. Then
\begin{equation}
  \E\left[\mathcal{Q}(\vDelta)\right]
  =
  \frac{1}{2}\tr\!\left(\rmH\,\E\left[\vDelta\vDelta^\top\right]\right)
  =
  \frac{\eta^2}{2}\tr(\rmH\rmM).
  \label{eq:second-moment-boundary}
\end{equation}
This blind benchmark intentionally ignores higher distributional details and
compares only how a fixed norm budget is exposed to unknown curvature.

To define the curvature-agnostic benchmark, compare exposures under a fixed trace
scale for retention curvature:
\begin{equation}
  \mathcal{H}_{\bar\lambda}
  \triangleq
  \left\{\rmH\succeq0:\tr(\rmH)=d\bar\lambda\right\},
  \qquad
  \mathcal{M}
  \triangleq
  \left\{\rmM\succeq0:\tr(\rmM)=\norm{\vg}^{2}\right\}.
  \label{eq:hblind-sets}
\end{equation}
Here $\bar\lambda$ denotes the unknown average curvature scale shared by the
benchmark class, and $\mathcal{M}$ is the norm-matched exposure budget
$\E\left[\norm{\rvx}^{2}\right]=\norm{\vg}^{2}$. If $\rmH$ is otherwise
unknown, the worst case places its curvature trace on the most exposed direction
of $\rmM$, so the safest target is the exposure with no preferred direction.

\begin{theorem}[Curvature-Agnostic Isotropic Benchmark]
\label{thm:hblind-isotropization}
Fix $\bar\lambda>0$ and $\vg\ne\vzero$. For any
$\rmM\in\mathcal{M}$, let $\lambda_{\max}(\rmM)$ denote its largest eigenvalue.
Then
\begingroup
\small
\begin{equation}
  \begin{alignedat}{2}
  \sup_{\rmH\in\mathcal{H}_{\bar\lambda}}
  \frac{\eta^2}{2}\tr(\rmH \rmM)
  &=
  \frac{\eta^2}{2}\,d\bar\lambda\,\lambda_{\max}(\rmM),
  \qquad&
  \inf_{\rmM\in\mathcal{M}}\ 
  \sup_{\rmH\in\mathcal{H}_{\bar\lambda}}
  \frac{\eta^2}{2}\tr(\rmH \rmM)
  &=
  \frac{\eta^2}{2}\bar\lambda\norm{\vg}^{2}.
  \end{alignedat}
  \label{eq:hblind-worst-exposure}
\end{equation}
\endgroup
The unique second moment attaining the second value is
$\rmM^\star=(\norm{\vg}^{2}/d)\rmI$.
\end{theorem}

Theorem~\ref{thm:hblind-isotropization} identifies the retention-only endpoint.
Under no curvature information, the minimax exposure is isotropic. This target
benchmarks exposure after the incoming direction has been fixed; it does not
prescribe a new plasticity objective. Norm-matched ZO then sits between FO's
rank-one exposure and this isotropic endpoint.

\begin{corollary}[Norm-Matched ZO Exposure Interpolation]
\label{cor:zo-exposure-interpolation}
Let $\tau=d/(q+d+1)$,
$\rmM^{\text{FO}}\triangleq\vg\vg^\top$, and
$\rmM^\star=(\norm{\vg}^{2}/d)\rmI$. For the norm-matched ZO shape in
\eqref{eq:shape-subcases},
\begin{equation}
  \rmM_q^{\text{ZO}}
  \triangleq
  \E\left[(\rmP\vg)(\rmP\vg)^\top\right]
  =
  (1-\tau)\vg\vg^\top
  +
  \tau\frac{\norm{\vg}^{2}}{d}\rmI
  =
  (1-\tau)\rmM^{\text{FO}}+\tau\rmM^\star .
  \label{eq:zo-exposure-interpolation}
\end{equation}
Moreover, with
$\mathcal{R}_{\bar\lambda}(\rmM)
\triangleq
\sup_{\rmH\in\mathcal{H}_{\bar\lambda}}
\frac{\eta^2}{2}\tr(\rmH\rmM)$,
\begin{equation}
  \mathcal{R}_{\bar\lambda}\!\left(\rmM_q^{\text{ZO}}\right)
  -
  \mathcal{R}_{\bar\lambda}\!\left(\rmM^\star\right)
  =
  (1-\tau)
  \left[
  \mathcal{R}_{\bar\lambda}\!\left(\rmM^{\text{FO}}\right)
  -
  \mathcal{R}_{\bar\lambda}\!\left(\rmM^\star\right)
  \right].
  \label{eq:zo-exposure-gap-closing}
\end{equation}
\end{corollary}

\begin{figure}[t]
  \vspace{-8mm}
  \centering
  \includegraphics[width=\linewidth]{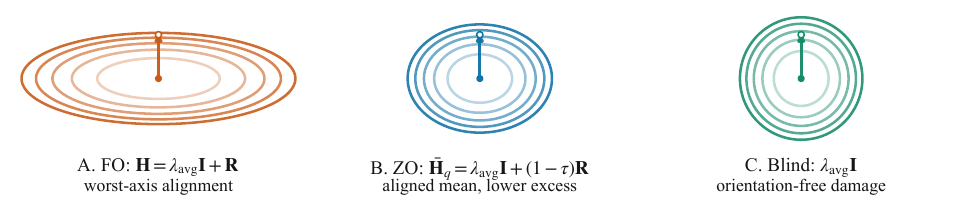}
  \vspace{-6mm}
  \caption{
  \textbf{Effective-curvature view of curvature-agnostic ZO shaping.}
  All panels use the same quadratic-damage levels; darker contours indicate
  larger damage, and the arrow marks the shared incoming direction $\vg$. The
  contours show the curvature read by the shaped adaptation: FO reads
  $\rmH=\lamavg\rmI+\rmR$, norm-matched ZO reads
  $\bar{\rmH}_{q}=\E[\rmP^\top\rmH\rmP]=\lamavg\rmI+(1-\tau)\rmR$, and the blind
  endpoint reads $\lamavg\rmI$. ZO therefore keeps the mean direction aligned
  while contracting anisotropic curvature toward the isotropic benchmark.
  }
  \label{fig:zo-shaping-mechanism}
  \vspace{-2mm}
\end{figure}

Corollary~\ref{cor:zo-exposure-interpolation} supplies the curvature-agnostic
interpretation. FO has rank-one exposure, the blind-optimal benchmark is
isotropic, and norm-matched ZO closes exactly a $\tau$-fraction of the
FO-to-optimum exposure gap, leaving
$1-\tau=(q+1)/(q+d+1)$. This benchmark explains why randomized shaping is
principled when $\rmH$ is unknown, but it does not replace the fixed-$\rmH$
forgetting identity in Theorem~\ref{thm:direct-damage}, which governs the actual
FO--ZO gap for a given retention geometry. Figure~\ref{fig:zo-shaping-mechanism}
visualizes the same mechanism in the equivalent shaped-curvature contour view.

\section{From Theory to \ours{}: Exact-Gradient Randomized Shaping}
\label{sec:method}

The theory isolates the transferable part of low-query ZO: after the calibration
in Lemmas~\ref{lem:finite-smoothing} and \ref{lem:mean-norm}, the
retention-relevant operation is a randomized shape applied to a fixed incoming
direction. \ours{} (\oursfull{}) transfers this operation to exact-gradient
adaptation. It computes the incoming FO gradient $\vg$, replaces it by a
calibrated shaped vector $\widetilde{\vg}$, and lets the same base optimizer
execute the adaptation. Thus \ours{} sits between FO and practical
finite-difference ZO. Relative to FO, it changes second-moment exposure as in
\eqref{eq:zo-exposure-interpolation} under a matched expected squared norm.
Relative to finite-difference ZO, it keeps the exact incoming signal, removes
finite-$\mu$ bias from \eqref{eq:finite-mu-bridge}, and replaces raw norm
inflation with explicit norm matching from \eqref{eq:norm-matched-calibration}.

\ours{} uses the same shape blockwise because global low-query shaping has
dimension-dependent deviation (Proposition~\ref{prop:finite-query-deviation}).
For blocks of size $d_b$, $q$ directions per block give $Bq$ local shaping
samples after one backward pass and no function-value queries. The block theory
keeps cross-block curvature in the mean term and treats omitted couplings as a
perturbation when small (Theorem~\ref{thm:rise-blockwise-mean};
Proposition~\ref{prop:block-coupling}). Thus \ours{} acts as a
stability--plasticity wrapper, importing ZO-style curvature mixing into
exact-gradient adaptation at a norm-matched mean-shrinkage cost relative to FO.

For a block partition $\vg=(\vg_1,\ldots,\vg_B)$, draw
$\rvz_{b,1},\ldots,\rvz_{b,q}\overset{\text{i.i.d.}}{\sim}\mathcal{N}(\vzero,\rmI_b)$
and define, for each $b$,
\begin{equation}
  \rmZ_b
  \triangleq
  \frac{1}{q}\sum_{i=1}^{q}\rvz_{b,i}\rvz_{b,i}^{\top},
  \qquad
  \kappa_b\triangleq\frac{q+d_b+1}{q}, \qquad
  \rmP_b
  \triangleq
  \kappa_b^{-1/2}\rmZ_b,
  \qquad
  \widetilde{\vg}_b
  \triangleq
  \rmP_b\vg_b .
  \label{eq:rise-block}
\end{equation}
By Lemma~\ref{lem:mean-norm}, each block is raw-aligned before normalization and
norm-matched after normalization:
$\E[\rmZ_b\vg_b]=\vg_b$,
$\E[\norm{\rmP_b\vg_b}^{2}]=\norm{\vg_b}^{2}$, and
$\E[\rmP_b\vg_b]=\kappa_b^{-1/2}\vg_b$. The shaped blocks replace the raw
gradient inside the same FO optimizer. Blocks may be tensors, layers, or optimizer groups; Appendix~\ref{app:rise-block-theory}
defines the blockwise scores used to interpret ablations.

\begin{algorithm}[t]
\DontPrintSemicolon
\caption{\ours{} (\oursfull{})}\label{alg:rise}
\KwIn{Exact gradient $\vg$, block partition $\{\vg_b\}_{b=1}^{B}$, directions per block $q$, base optimizer \textsc{Opt}}
\For{each block $b=1,\ldots,B$}{
  Draw $\rvz_{b,1},\ldots,\rvz_{b,q}\overset{\text{i.i.d.}}{\sim}\mathcal{N}(\vzero,\rmI_b)$\;
  $\rmZ_b \leftarrow q^{-1}\sum_{i=1}^{q}\rvz_{b,i}\rvz_{b,i}^{\top}$ and $\kappa_b\leftarrow(q+d_b+1)/q$\;
  $\widetilde{\vg}_b \leftarrow \kappa_b^{-1/2}\rmZ_b \vg_b$\;
}
Concatenate $\widetilde{\vg} \leftarrow (\widetilde{\vg}_1,\ldots,\widetilde{\vg}_B)$\;
Pass $\widetilde{\vg}$ to \textsc{Opt} in place of $\vg$\;
\end{algorithm}

\section{Experiments}
\label{sec:experiments}

The main experiments test whether the local shaping mechanism survives inside
continual-learning pipelines. Prior forward-pass and ZO work shows that ZO can
reduce forgetting in ZeroFlow \citep{feng2025zeroflow}, forward-only continual
adaptation \citep{chen2025forwardonly}, and ZO-FC \citep{yu2025more}; here the
question is whether the retention-relevant ZO shape can be applied to exact
gradients without finite-difference plasticity loss. We evaluate ViT-B/16 on
CIFAR100 \citep{krizhevsky2009learning}, ImageNet-R \citep{hendrycks2021many},
an ImageNet-derived rendition benchmark \citep{russakovsky2015imagenet}, and
DomainNet \citep{peng2019moment}, and T5 on task orders from the
text-classification CL benchmark \citep{zhang2015character}, GLUE
\citep{wang2018glue}, and SuperGLUE \citep{wang2019superglue}. Metrics are Avg,
Last, and Fgt; full settings and quadratic-sandbox evidence are in
Appendix~\ref{app:extended-experiments}.

\paragraph{Vision results.}
In vision, \ours{} keeps the ZO-FC classifier/head design and replaces
representation-side finite-difference ZO with exact-gradient norm-matched
shaping. Table~\ref{tab:rise-fc-results} shows that ZO-FC+\ours{} improves Last
over ZO-FC by $+1.39$, $+0.94$, and $+0.39$ points on CIFAR100, ImageNet-R, and
DomainNet, while reducing Fgt by $1.01$, $1.72$, and $0.28$ points. Higher Last
indicates that exact gradients recover plasticity lost to gradient estimation;
lower Fgt at comparable or higher Avg indicates that the ZO-derived shape still
reduces retention exposure. Strong FO baselines and low-forgetting but
low-plasticity ZO rows rule out one-axis dominance; the gain is the paired
balance over the closest ZO-FC pipeline.

\begin{table}[t]
  \centering
  \setlength{\tabcolsep}{7pt}
  \caption{Vision continual-learning results (\%).}
  \begin{tabular}{@{}lccccccccc@{}}
    \toprule
    & \multicolumn{3}{c}{\textbf{CIFAR100 Inc-5}}
    & \multicolumn{3}{c}{\textbf{ImageNet-R Inc-5}}
    & \multicolumn{3}{c}{\textbf{DomainNet Inc-69}} \\
    \cmidrule(lr){2-4}\cmidrule(lr){5-7}\cmidrule(l){8-10}
    \textbf{Method}
    & avg$\uparrow$ & last$\uparrow$ & fgt$\downarrow$
    & avg$\uparrow$ & last$\uparrow$ & fgt$\downarrow$
    & avg$\uparrow$ & last$\uparrow$ & fgt$\downarrow$ \\
    \midrule
    \multicolumn{10}{c}{\textbf{FO Optimization}} \\
    \midrule
    Learnable Cls. \citep{hou2019learning}
    & 85.61 & 78.33 & 5.17
    & 60.60 & 52.12 & \underline{3.98}
    & 75.85 & 69.88 & \textbf{3.70} \\
    Adapter + Cls. \citep{yu2025more}
    & 88.41 & 83.64 & 9.58
    & \underline{74.71} & \underline{69.62} & 8.49
    & \textbf{78.67} & \textbf{72.85} & 4.64 \\
    APER Adapter \citep{zhou2025revisiting}
    & 87.90 & 81.68 & 6.01
    & 71.52 & 63.42 & 7.48
    & 71.24 & 64.93 & 6.72 \\
    EASE \citep{zhou2024expandable}
    & \textbf{91.13} & \textbf{85.45} & 7.05
    & \textbf{78.78} & \textbf{71.93} & 7.54
    & 75.91 & 70.24 & 8.54 \\
    \midrule
    \multicolumn{10}{c}{\textbf{ZO Optimization}} \\
    \midrule
    Learnable Cls.
    & 62.85 & 43.55 & \textbf{3.99}
    & 25.60 & 17.08 & 9.18
    & 52.70 & 46.21 & 7.37 \\
    Adapter + Cls.
    & 79.45 & 70.46 & 8.72
    & 34.87 & 27.33 & 6.36
    & 48.24 & 44.29 & 4.56 \\
    APER Adapter
    & 87.70 & 81.49 & 6.07
    & 62.60 & 54.50 & 8.55
    & 62.45 & 56.67 & 6.89 \\
    EASE
    & 87.32 & 81.07 & 6.87
    & 62.65 & 54.47 & 9.84
    & 63.10 & 56.80 & 7.22 \\
    \midrule
    ZO-FC \citep{yu2025more}
    & 88.39 & 83.34 & 5.26
    & 72.01 & 66.63 & 4.40
    & 77.17 & 71.05 & 4.19 \\
    ZO-FC+\ours{}
    & \underline{89.75} & \underline{84.73} & \underline{4.25}
    & 72.05 & 67.57 & \textbf{2.68}
    & \underline{77.57} & \underline{71.44} & \underline{3.91} \\
    \bottomrule
  \end{tabular}
  \label{tab:rise-fc-results}
\end{table}

\paragraph{Language-model transfer.}
Table~\ref{tab:additional-cl-benchmarks} reports language transfer on SeqLoRA
and O-LoRA. On SeqLoRA, \ours{} keeps Last close to FO while cutting FO
forgetting from $35.57$ to $7.01$, from $37.01$ to $6.58$, and from $50.62$ to
$17.77$. O-LoRA starts from a lower-forgetting FO baseline, yet \ours{} still
lowers Fgt on all orders while maintaining or improving Avg. The contrast
matches the mechanism: when FO forgets heavily, the shaped exact gradient can
reduce retention exposure without losing the final-task signal; when FO is
already stable, gains are smaller but consistent because the exact gradient
carries the incoming signal while randomized shape changes its retention
exposure.

\begin{table}[t]
  \centering
  \small
  \setlength{\tabcolsep}{0pt}
  \newcommand{\methodcell}[1]{\makebox[0.48in][c]{\shortstack{#1}}}
  \newcommand{\metriccell}[1]{\makebox[0.47in][c]{#1}}
  \newcommand{\variantcell}[1]{\makebox[0.30in][c]{#1}}
  \newcommand{\benchmarkcell}[1]{\makebox[1.42in][c]{\textbf{\shortstack{#1}}}}
  \newcommand{\resultstd}[1]{\makebox[0pt][l]{\hspace{0.05em}\raisebox{-0.45ex}{\scalebox{0.48}{\ensuremath{\pm#1}}}}}
  \newcommand{\mockstd}{\resultstd{0.00}}
  \caption{Additional language-model continual-learning results (\%). Values
  report the mean with standard deviation shown after $\pm$ over repeated runs
  under the same task order.}
  \begin{tabular}{@{}c@{\hspace{2pt}}c@{\hspace{3pt}}ccc@{\hspace{6pt}}ccc@{\hspace{6pt}}ccc@{}}
    \toprule
    & & \multicolumn{3}{c@{\hspace{6pt}}}{\benchmarkcell{Standard CL Benchmark \\ Order 1}}
    & \multicolumn{3}{c@{\hspace{6pt}}}{\benchmarkcell{Standard CL Benchmark \\ Order 2}}
    & \multicolumn{3}{c@{}}{\benchmarkcell{Large Number of Tasks \\ Order 3}} \\
    \cmidrule(lr){3-5}\cmidrule(lr){6-8}\cmidrule(l){9-11}
    \methodcell{\textbf{Method}} & \variantcell{\textbf{Type}}
    & \metriccell{Avg$\uparrow$} & \metriccell{Last$\uparrow$} & \metriccell{Fgt$\downarrow$}
    & \metriccell{Avg$\uparrow$} & \metriccell{Last$\uparrow$} & \metriccell{Fgt$\downarrow$}
    & \metriccell{Avg$\uparrow$} & \metriccell{Last$\uparrow$} & \metriccell{Fgt$\downarrow$} \\
    \midrule
    \multirow{3}{*}{\methodcell{SeqLoRA\\[-1pt]{\scriptsize\citep{hu2022lora}}}} & FO
    & \underline{53.33}\resultstd{1.16} & \textbf{89.99}\resultstd{0.40} & 35.57\resultstd{1.50} & \underline{52.13}\resultstd{4.89} & \textbf{73.23}\resultstd{0.61} & 37.01\resultstd{6.69} & \underline{27.62}\resultstd{11.26} & 52.43\resultstd{3.81} & 50.62\resultstd{12.43} \\
    & ZO
    & 46.24\resultstd{3.02} & 62.05\resultstd{5.49} & \textbf{3.67}\resultstd{0.62} & 42.07\resultstd{3.62} & 57.51\resultstd{1.31} & \underline{7.36}\resultstd{1.31} & 25.60\resultstd{8.24} & \underline{53.61}\resultstd{10.77} & \textbf{15.83}\resultstd{6.13} \\
    & \ours{}
    & \textbf{74.85}\resultstd{1.95} & \underline{89.46}\resultstd{0.09} & \underline{7.01}\resultstd{2.21} & \textbf{75.15}\resultstd{0.67} & \underline{73.16}\resultstd{0.23} & \textbf{6.58}\resultstd{1.45} & \textbf{62.65}\resultstd{5.13} & \textbf{56.66}\resultstd{1.42} & \underline{17.77}\resultstd{5.16} \\
    \midrule
    \multirow{3}{*}{\methodcell{O-LoRA\\[-1pt]{\scriptsize\citep{wang2023orthogonal}}}} & FO
    & \underline{77.19}\resultstd{0.28} & \textbf{89.04}\resultstd{0.47} & 2.83\resultstd{0.02} & \underline{76.61}\resultstd{0.62} & \textbf{71.92}\resultstd{0.19} & 2.37\resultstd{0.32} & \underline{72.72}\resultstd{1.61} & \underline{54.44}\resultstd{2.09} & 3.80\resultstd{1.44} \\
    & ZO
    & 62.42\resultstd{1.47} & 61.70\resultstd{3.18} & \textbf{0.98}\resultstd{1.16} & 60.38\resultstd{1.32} & 61.72\resultstd{0.86} & \underline{2.09}\resultstd{2.59} & 56.48\resultstd{2.61} & 52.19\resultstd{6.54} & \underline{3.76}\resultstd{2.57} \\
    & \ours{}
    & \textbf{77.24}\resultstd{0.30} & \underline{88.81}\resultstd{0.37} & \underline{1.43}\resultstd{0.37} & \textbf{76.90}\resultstd{0.18} & \underline{71.85}\resultstd{0.26} & \textbf{1.87}\resultstd{0.05} & \textbf{74.06}\resultstd{0.39} & \textbf{54.55}\resultstd{1.27} & \textbf{3.16}\resultstd{0.48} \\
    \bottomrule
  \end{tabular}
  \label{tab:additional-cl-benchmarks}
\end{table}

Together, the domains support the same conditional pattern: randomized shaping
helps when it lowers shape-sensitive retention exposure without collapsing the
incoming plasticity signal, and has less room when the base method already
forgets little. The raw-ZO rows show why exact-gradient transfer matters: low
forgetting can coexist with poor Last, so retention shape must be separated from
finite-difference plasticity loss before attributing gains to ZO-style curvature
mixing.

\section{Conclusion}
\label{sec:conclusion}

Low-query ZO retention is calibrated randomized shaping: raw ZO supplies the
mean-aligned signal, norm matching fixes squared norm with explicit mean
shrinkage, and the FO--ZO damage gap becomes a query-dependent fraction of
directional curvature excess. \ours{} transfers this shape to exact gradients as
a blockwise wrapper, yielding a local second-moment stability--plasticity
tradeoff whose global trajectory and selection questions are handled by the
appendix controls and blockwise analyses.

\bibliographystyle{plainnat}
\bibliography{workspace/reference}

\appendix
\section{Related Work and Positioning}
\label{sec:related}

\paragraph{Continual Learning and Retention Geometry.}
Continual-learning methods have long controlled forgetting by restricting how a
new-task adaptation may move the model. Learning without Forgetting preserves
old behavior through distillation \citep{li2016learning}; EWC penalizes movement
in Fisher-important parameters \citep{kirkpatrick2017overcoming}; iCaRL stores
exemplars for replay \citep{rebuffi2017icarl}; GEM and A-GEM impose
past-gradient constraints \citep{lopezpaz2017gradient,chaudhry2019efficient};
and OGD and GPM project updates away from past-task subspaces
\citep{farajtabar2020orthogonal,saha2021gradient}. More recent geometric work
makes the same stability concern curvature-aware: AlterSGD searches for flat
continual-learning minima \citep{huang2021altersgd}, SAM penalizes sharp
solutions \citep{foret2021sharpness}, C-Flat adapts flatness control to
continual learning \citep{bian2024cflat}, and Hessian-eigenvalue analysis links
forgetting to high-curvature directions \citep{kong2024overcoming}. These
methods use explicit retention information to change the objective, gradient, or
training rule. Our analysis isolates a smaller local object that exists before
such retention channels are added: once the incoming-loss direction is fixed,
historical loss only sees the induced adaptation displacement and the retention
curvature exposed by that displacement.

\paragraph{Pretrained-Model and PEFT Continual Adaptation.}
The empirical baselines in this paper sit on a newer line that adapts
pretrained representations rather than training continual learners from
scratch. Unified classifier rebalancing is an early class-incremental
classifier baseline \citep{hou2019learning}. LoRA supplies the low-rank adapter
parameterization used by many language-model continual-learning pipelines
\citep{hu2022lora}, while O-LoRA adds orthogonal subspace constraints for
language-model continual learning \citep{wang2023orthogonal}. For vision
continual learning with pretrained models, EASE expands subspace ensembles
\citep{zhou2024expandable}, and APER emphasizes adaptive pretrained-model
class-incremental learning \citep{zhou2025revisiting}. These methods define
strong host settings and baseline rows, but they do not by themselves explain
why a ZO-shaped adaptation should expose less retention curvature than an
ordinary FO displacement under the same norm budget.

\paragraph{Zeroth-Order Optimization and Forward-Pass Adaptation.}
Classical ZO optimization estimates descent directions from function values,
from bandit finite differences \citep{flaxman2005online} to randomized
gradient-free smoothing \citep{nesterov2017random}. In modern language-model
fine-tuning, MeZO shows that large models can be adapted with forward passes
alone \citep{malladi2023fine}, and the ZO-LLM benchmark broadens this evidence
across optimizers, model families, and tuning schemes \citep{zhang2024revisiting}.
Subsequent variants reduce the estimation burden by selecting sparse parameter
subsets \citep{liu2024sparsemezo}, using low-rank ZO structures
\citep{chen2025lozo}, adapting layerwise ZO update scales
\citep{tan2025harmony}, or combining ZO and FO optimization
\citep{sugiura2025elasticzo}. These works establish the feasibility and systems
value of forward-pass fine-tuning. Their main object, however, is still gradient
estimation or memory-efficient optimization, not the retention curvature exposed
after an incoming gradient has already been fixed.

\paragraph{Forward-Pass and ZO Continual Learning.}
The closest empirical continual-learning work moves from memory-efficient ZO to
forgetting itself. ZeroFlow benchmarks gradient-free algorithms for continual
learning and shows that forward passes can be sufficient in several forgetting
settings \citep{feng2025zeroflow}; Forward-Only Continual Learning explores
gradient-free adaptation for pretrained models \citep{chen2025forwardonly}; and
the ZO-FC study shows that ZO optimization can reduce forgetting beyond memory
savings, identifies a stability--plasticity tradeoff, and optimizes an adapter
with ZO while keeping the classifier/head under FO gradients \citep{yu2025more}.
These papers establish the phenomenon that motivates this work. The open
mechanistic question is what remains after finite-smoothing error, raw-ZO mean
alignment, raw norm inflation, norm-matched mean shrinkage, and expected squared
adaptation norm are separated. We answer that question with a direct
norm-matched FO--ZO forgetting-gap identity, and \ours{} then transfers the same
shape mechanism to exact gradients.

The remaining appendix gives the finite-query account
(Appendix~\ref{app:practical-zo-gap}), the blockwise \ours{} analysis
(Appendix~\ref{app:rise-block-theory}), the empirical settings and controls
(Appendix~\ref{app:extended-experiments}), and the proofs
(Appendix~\ref{app:proofs}).

\section{Practical Finite-Query ZO Gap}
\label{app:practical-zo-gap}

The mean zero-smoothing mechanism leaves one implementation question: whether a
single finite-difference batch realizes the same sign and scale. The resulting
realization-level account keeps the two perturbations separate.
Finite-query randomness is a centered sampling deviation around the mean gap,
while finite smoothing is an estimator remainder that perturbs the quadratic
damage.

\begin{proposition}[One-Batch Deviation around the Mean Gap]
\label{prop:finite-query-deviation}
Let $\rmZ=q^{-1}\sum_{r=1}^{q}\rvz_r\rvz_r^\top$ with
$\rvz_r\overset{\text{i.i.d.}}{\sim}\mathcal{N}(\vzero,\rmI)$,
$\rmP=\kappa^{-1/2}\rmZ$, $\kappa=(q+d+1)/q$, and $\tau=d/(q+d+1)$.
Assume $\rmH\succeq0$ and let $\lambda_{\max}$ denote its largest eigenvalue.
Let
$G_q=(\eta^2/2)\tau\norm{\vg}^{2}(\lamdir-\lamavg)$ be the mean
forgetting-reduction signal in \eqref{eq:mean-gap-signal}, where
$\lamdir=\vg^\top\rmH\vg/\norm{\vg}^{2}$ and
$\lamavg=\tr(\rmH)/d$. For
$\delta\in(0,1)$, define
\begin{equation}
  \psi_q(\delta)
  \triangleq
  \sqrt{\frac{d+\log(1/\delta)}{q}}
  +
  \frac{d+\log(1/\delta)+1}{q}.
  \label{eq:finite-query-fluctuation-scale}
\end{equation}
There is a universal constant $C>0$ such that, with probability at least
$1-\delta$,
\begin{equation}
  \left|
  \left(
  \mathcal{Q}^{\text{FO}}
  -
  \mathcal{Q}_q^{\text{ZO}}
  \right)
  -
  G_q
  \right|
  =
  \left|
  \mathcal{Q}_q^{\text{ZO}}
  -
  \E\left[\mathcal{Q}_q^{\text{ZO}}\right]
  \right|
  \le
  C\frac{\eta^2}{2}\norm{\vg}^{2}
  \lambda_{\max}
  \left(\psi_q(\delta)+\psi_q(\delta)^2\right).
  \label{eq:finite-query-scalar-bound}
\end{equation}
\end{proposition}

Proposition~\ref{prop:finite-query-deviation} turns a sampled ZO batch into a
signal--deviation comparison. The deterministic signal is $G_q$, while the
right-hand side of \eqref{eq:finite-query-scalar-bound} is the local
sample-covariance scale multiplied by the largest retention curvature. A
positive mean benefit is visible in one batch only when the signal dominates
this deviation; a negative mean gap is visible by the same logic. The query
count therefore has two roles: smaller $q$ increases the mean mixing weight
$\tau$, but also increases the realization scale $\psi_q(\delta)$.

\begin{corollary}[Finite-Smoothing Perturbation to the Realized Gap]
\label{cor:finite-smoothing-damage}
Assume the conditions of Lemma~\ref{lem:finite-smoothing}. Also assume
$\rmH\succeq0$ and let $\lambda_{\max}$ be its largest eigenvalue. Define
the norm-matched finite-difference direction
$\widehat{\rvx}_{\mu}\triangleq\kappa^{-1/2}\widehat{\vg}_{\mu}$ and its
quadratic damage
$\widehat{\mathcal{Q}}_{q,\mu}^{\text{ZO}}
\triangleq\mathcal{Q}(-\eta\widehat{\rvx}_{\mu})$. Then the realized-gap
accounting in \eqref{eq:practical-finite-mu-accounting} holds. Moreover, the
same finite-smoothing residual controls both the damage perturbation and the
small mismatch from exact zero-smoothing norm matching. For a universal constant
$C>0$,
\begin{equation}
  \E\left[
  \left|
  \widehat{\mathcal{Q}}_{q,\mu}^{\text{ZO}}
  -
  \mathcal{Q}_q^{\text{ZO}}
  \right|
  \right]
  \le
  C\frac{\eta^2}{2}\lambda_{\max}
  \left[
  \kappa^{-1/2}L\mu d^{3/2}\norm{\vg}
  +
  \kappa^{-1}L^2\mu^2 d^3
  \right].
  \label{eq:finite-smoothing-damage}
\end{equation}
The expected squared norm obeys
\begin{equation}
  \left|
  \E\left[\norm{\widehat{\rvx}_{\mu}}^{2}\right]
  -
  \norm{\vg}^{2}
  \right|
  \le
  C
  \left[
  \kappa^{-1/2}L\mu d^{3/2}\norm{\vg}
  +
  \kappa^{-1}L^2\mu^2 d^3
  \right].
  \label{eq:finite-smoothing-norm-mismatch}
\end{equation}
\end{corollary}

The finite-difference residual enters the quadratic damage in two ways: through
a cross term with the zero-smoothing shaped direction and through the residual
quadratic term. The expected-norm bound in Lemma~\ref{lem:finite-smoothing}
justifies the estimator-level reduction, while the second-moment bound
$\E[\norm{\vr_{\mu}}^2]$ controls the damage-level perturbation through
Cauchy--Schwarz and the residual quadratic term. Thus
\eqref{eq:practical-finite-mu-accounting} realizes the main-text mean mechanism
when $G_q$ dominates both the finite-query deviation and the finite-smoothing
perturbation at the relevant scale. Equation~\eqref{eq:finite-smoothing-norm-mismatch}
also makes explicit that norm matching is exact for the zero-smoothing shape and
perturbed, at the same order, for a finite-difference batch.

\begin{corollary}[Realized Gap Sign Certificate]
\label{cor:realized-gap-sign}
Under the conditions of Proposition~\ref{prop:finite-query-deviation} and
Corollary~\ref{cor:finite-smoothing-damage}, define
\begin{align}
  B_q(\delta)
  &\triangleq
  C\frac{\eta^2}{2}\norm{\vg}^{2}
  \lambda_{\max}
  \left(\psi_q(\delta)+\psi_q(\delta)^2\right),\\
  B_{\mu}
  &\triangleq
  C\frac{\eta^2}{2}\lambda_{\max}
  \left[
  \kappa^{-1/2}L\mu d^{3/2}\norm{\vg}
  +
  \kappa^{-1}L^2\mu^2 d^3
  \right],
\end{align}
where $C$ is a universal constant large enough for the two preceding results.
For any $\delta_q,\delta_\mu\in(0,1)$, with probability at least
$1-\delta_q-\delta_\mu$,
\begin{equation}
  \left|
  \left(
  \mathcal{Q}^{\text{FO}}
  -
  \widehat{\mathcal{Q}}_{q,\mu}^{\text{ZO}}
  \right)
  -
  G_q
  \right|
  \le
  B_q(\delta_q)+\frac{B_\mu}{\delta_\mu}.
  \label{eq:realized-gap-certificate}
\end{equation}
Consequently, if
$|G_q|>B_q(\delta_q)+B_\mu/\delta_\mu$, the practical realized gap has the same
sign as the mean signal $G_q$. In the zero-smoothing setting, set $B_\mu=0$.
\end{corollary}

If the two function values are observed with additive zero-mean scalar noise,
the estimator gains an additional residual. For independent noises with
conditional variance at most $\sigma_f^2$ per function value, the residual
$\vn_{\mu}=q^{-1}\sum_{r=1}^{q}
((\xi_{r,+}-\xi_{r,-})/(2\mu))\rvz_r$ satisfies
$\E[\norm{\vn_{\mu}}^2]\le \sigma_f^2 d/(2\mu^2q)$. The same perturbation
calculation applies after replacing $\vr_{\mu}$ by
$\vr_{\mu}+\vn_{\mu}$. This is the reason practical ZO experiments either use
common mini-batches/common random numbers for the two function values or report
the additional variance induced by stochastic function-value observations.

\section{Blockwise Support Theory for \ours{}}
\label{app:rise-block-theory}

\ours{} applies the calibrated ZO shape independently inside parameter blocks,
so the global identity from Sec.~\ref{sec:theory} is localized rather than
reused verbatim. The blockwise support does not claim that vanilla ZO is
intrinsically blockwise. A full-$\rmH$ oracle could exploit intra-block and
inter-block curvature simultaneously; a blockwise informed rule would keep the
within-block matrices $\rmH_{bb}$ and ignore cross-block couplings. \ours{} uses
weaker information by shaping natural parameter blocks independently. The local
calculation identifies how much of the global identity survives this
localization and which term records the cost of ignoring cross-block curvature.

Partition the parameter into $B$ blocks and write the retention curvature in
blocks $\rmH_{bc}$. Let block $b$ have dimension $d_b$ and identity $\rmI_b$.
For each block, draw
$\rvz_{b,1},\ldots,\rvz_{b,q_b}\overset{\text{i.i.d.}}{\sim}
\mathcal{N}(\vzero,\rmI_b)$ independently across blocks and define
\begin{equation}
  \rmZ_b
  \triangleq
  \frac{1}{q_b}\sum_{i=1}^{q_b}\rvz_{b,i}\rvz_{b,i}^{\top},
  \qquad
  \kappa_b\triangleq\frac{q_b+d_b+1}{q_b},
  \qquad
  \rmP_b\triangleq\kappa_b^{-1/2}\rmZ_b .
  \label{eq:app-rise-block-shape}
\end{equation}
Let $\rmP_{\text{blk}}\triangleq\operatorname{blkdiag}(\rmP_1,\ldots,\rmP_B)$,
$a_b\triangleq\kappa_b^{-1/2}$, and
$\tau_b\triangleq d_b/(q_b+d_b+1)$.

\begin{theorem}[Blockwise Expected Curvature and Gap for \ours{}]
\label{thm:rise-blockwise-gap}
For the blockwise \ours{} shape $\rmP_{\text{blk}}$ in
\eqref{eq:app-rise-block-shape}, any fixed symmetric retention curvature
$\rmH$ satisfies
\begin{equation}
  \left(\E\left[\rmP_{\text{blk}}^\top\rmH\rmP_{\text{blk}}\right]\right)_{bb}
  =
  (1-\tau_b)\rmH_{bb}
  +
  \tau_b\lamavg_b\rmI_b,
  \qquad
  \left(\E\left[\rmP_{\text{blk}}^\top\rmH\rmP_{\text{blk}}\right]\right)_{bc}
  =
  a_ba_c\rmH_{bc}\quad(b\ne c),
  \label{eq:rise-blockwise-curvature}
\end{equation}
where $\lamavg_b\triangleq\tr(\rmH_{bb})/d_b$. Consequently, with
$\mathcal{Q}^{\text{\ours}}_{\text{blk}}
\triangleq\mathcal{Q}(-\eta\rmP_{\text{blk}}\vg)$,
\begin{equation}
  \mathcal{Q}^{\text{FO}}
  -
  \E\left[\mathcal{Q}^{\text{\ours}}_{\text{blk}}\right]
  =
  \frac{\eta^2}{2}
  \left[
  \sum_{b=1}^{B}
  \tau_b\,
  \vg_b^\top(\rmH_{bb}-\lamavg_b\rmI_b)\vg_b
  +
  2\sum_{b<c}
  (1-a_ba_c)\vg_b^\top\rmH_{bc}\vg_c
  \right].
  \label{eq:rise-blockwise-gap}
\end{equation}
\end{theorem}

Theorem~\ref{thm:rise-blockwise-gap} is the formal connection between the
global mechanism and the blockwise wrapper. If $\rmH$ is block diagonal, the
gap is exactly the sum of the global FO--ZO identity applied inside each block.
If cross-block curvature is present, the second term in
\eqref{eq:rise-blockwise-gap} is the explicit price of blockwise localization:
it may help or hurt depending on the signed cross-block interaction
$\vg_b^\top\rmH_{bc}\vg_c$. This is why the blockwise evaluation uses the
blockwise mean-matched FO control $\rmA\vg$ and the covariance-matched control
in \eqref{eq:block-covariance-control}; otherwise shrinkage of cross-block mean
terms could be mistaken for a shaping effect.

\begin{proposition}[Blockwise One-Batch Deviation]
\label{prop:rise-blockwise-deviation}
Assume the setting of Theorem~\ref{thm:rise-blockwise-gap}. For
$\delta_b\in(0,1)$ define
\begin{equation}
  \psi_b(\delta_b)
  \triangleq
  \sqrt{\frac{d_b+\log(1/\delta_b)}{q_b}}
  +
  \frac{d_b+\log(1/\delta_b)+1}{q_b}.
  \label{eq:blockwise-fluctuation-scale}
\end{equation}
There is a universal constant $C>0$ such that, with probability at least
$1-\sum_{b=1}^{B}\delta_b$,
\begin{align}
  \left|
  \mathcal{Q}^{\text{\ours}}_{\text{blk}}
  -
  \E\left[\mathcal{Q}^{\text{\ours}}_{\text{blk}}\right]
  \right|
  &\le
  \frac{\eta^2}{2}
  \sum_{b=1}^{B}
  C\norm{\vg_b}^{2}
  \norm{\rmH_{bb}}_{\text{op}}
  \left(\psi_b(\delta_b)+\psi_b(\delta_b)^2\right)
  \nonumber\\
  &\quad+
  \eta^2
  \sum_{b<c}
  a_ba_c\norm{\vg_b}\norm{\vg_c}
  \norm{\rmH_{bc}}_{\text{op}}
  \left(\psi_b(\delta_b)+\psi_c(\delta_c)
  +\psi_b(\delta_b)\psi_c(\delta_c)\right).
  \label{eq:rise-blockwise-deviation}
\end{align}
\end{proposition}

Proposition~\ref{prop:rise-blockwise-deviation} explains the finite-query
advantage and its limitation. The within-block sampling scale depends on
$d_b$, not the full ambient dimension, so moderate blocks can make the
mean-curvature signal easier to realize with low $q_b$. The cross-block term
does not vanish automatically; it is controlled by the off-diagonal curvature
and by the shaped block norms. Thus blockwise shaping is a stable realization
of the global mechanism only when the within-block anisotropic signal is large
relative to both within-block sampling fluctuation and cross-block coupling.

The same blockwise object can be rewritten as a mean adaptation plus centered
random residual. This decomposition is useful for interpretation because it
separates deterministic mean-step damage from the expected cost of randomized
exposure.

\begin{theorem}[Expected Forgetting Decomposition for Blockwise Randomized Shaping]
\label{thm:rise-blockwise-mean}
Under the local quadratic forgetting model in \eqref{eq:quadratic-forgetting},
let a randomized adaptation decompose as $\vDelta=\vm+\veps$, where
$\vm=\E[\vDelta]$ and $\veps=(\veps_1,\ldots,\veps_B)$ is partitioned by
blocks. Assume
$\E[\veps_b]=\vzero$,
$\E[\veps_b\veps_c^\top]=\vzero$ for $b\ne c$, and
$\E[\veps_b\veps_b^\top]=\rmSigma_b$. With
$\rmSigma=\operatorname{blkdiag}(\rmSigma_1,\ldots,\rmSigma_B)$,
\begin{equation}
  \E\left[\mathcal{Q}(\vDelta)\right]
  =
  \frac{1}{2}\vm^\top\rmH\vm
  +
  \frac{1}{2}\tr(\rmH\rmSigma)
  =
  \frac{1}{2}\vm^\top\rmH\vm
  +
  \frac{1}{2}\sum_{b=1}^{B}\tr(\rmH_{bb}\rmSigma_b).
  \label{eq:rise-blockwise-mean}
\end{equation}
\end{theorem}

Theorem~\ref{thm:rise-blockwise-mean} is the expectation identity that matches
the blockwise wrapper. The first term is the damage of the mean adaptation and
may involve the full curvature $\rmH$, including cross-block couplings. This
term is why the empirical setup includes mean-scaled FO controls. After the
mean adaptation is controlled, the expected damage introduced by the centered
blockwise random residual depends only on $\rmH_{bb}$ and the block covariances
$\rmSigma_b$. Cross-block curvature is not irrelevant; it enters the stability
of one sampled realization. For a zero-mean blockwise residual,
\begin{equation}
  \mathcal{Q}(\veps)
  -
  \E\left[\mathcal{Q}(\veps)\right]
  =
  \frac{1}{2}\sum_{b=1}^{B}
  \left(
  \veps_b^\top\rmH_{bb}\veps_b
  -
  \tr(\rmH_{bb}\rmSigma_b)
  \right)
  +
  \sum_{b<c}\veps_b^\top\rmH_{bc}\veps_c .
  \label{eq:block-coupling-fluctuation}
\end{equation}
Thus the mean randomized damage is block-diagonal, while inter-block curvature
controls run-to-run fluctuation and the reliability of a single sampled shaped
adaptation.

The same decomposition gives a trace-level measure when a block is shaped
isotropically. If $\rmSigma_b=\alpha_b\rmI_b$, then
\begin{equation}
  \E\left[\mathcal{Q}(\veps)\right]
  =
  \frac{1}{2}\sum_{b=1}^{B}\alpha_b\tr(\rmH_{bb}),
  \qquad
  D_b\triangleq\frac{1}{d_b}\tr(\rmH_{bb}).
  \label{eq:block-damage-density}
\end{equation}
The damage density $D_b$ is the mean forgetting cost per unit isotropic
randomized exposure in block $b$. It is milder than a largest-eigenvalue score
and matches the average behavior of randomized shaping. Directional block
geometry gives a complementary score. For blocks with positive definite or
damped curvature, define
\begin{equation}
  R_b(\rho)
  \triangleq
  \vg_b^\top(\rmH_{bb}+\rho\rmI_b)^{-1}\vg_b,
  \qquad
  \rho\ge0 .
  \label{eq:block-flat-signal-score}
\end{equation}
Large $R_b(\rho)$ means the incoming signal in block $b$ can be carried through
directions that are flat for the historical loss. A block is therefore
forgetting-resistant not merely because $D_b$ is small, but because it combines
low average retention curvature, signal alignment with flat directions, and
weak coupling to the rest of the network. A shape-aware coupling score is
\begin{equation}
  C_b
  \triangleq
  \sum_{c\ne b}
  \norm{\rmSigma_b^{1/2}\rmH_{bc}\rmSigma_c^{1/2}}_{\text{F}},
  \label{eq:block-coupling-score}
\end{equation}
with the raw alternative
$\sum_{c\ne b}\norm{\rmH_{bc}}_{\text{F}}$ when the block covariance is not yet
specified. Blocks with modest $D_b$ but large $C_b$ can look safe on average
while showing unstable realized forgetting; these are fragile bridge blocks
rather than uniformly robust layers.

The distance from a full-$\rmH$ oracle is controlled by how large the omitted
couplings are relative to the block diagonal. Let
$\rmH_{\text{blk}}\triangleq\operatorname{blkdiag}(\rmH_{11},\ldots,\rmH_{BB})$,
$\rmE\triangleq\rmH-\rmH_{\text{blk}}$, and
\begin{equation}
  \epsilon_{\text{blk}}
  \triangleq
  \norm{
  \rmH_{\text{blk}}^{-1/2}\rmE\rmH_{\text{blk}}^{-1/2}
  }_{\text{op}} .
  \label{eq:block-coupling-coefficient}
\end{equation}

\begin{proposition}[Block-Coupling Perturbation]
\label{prop:block-coupling}
Let $\rmH_{\text{blk}}=\operatorname{blkdiag}(\rmH_{11},\ldots,\rmH_{BB})$,
$\rmE=\rmH-\rmH_{\text{blk}}$, and
$\epsilon_{\text{blk}}
=\norm{\rmH_{\text{blk}}^{-1/2}\rmE\rmH_{\text{blk}}^{-1/2}}_{\text{op}}$.
If $\rmH_{\text{blk}}\succ0$ and $\epsilon_{\text{blk}}<1$, then
\begin{equation}
  (1+\epsilon_{\text{blk}})^{-1}\rmH_{\text{blk}}^{-1}
  \preceq
  \rmH^{-1}
  \preceq
  (1-\epsilon_{\text{blk}})^{-1}\rmH_{\text{blk}}^{-1}.
  \label{eq:block-inverse-perturbation}
\end{equation}
\end{proposition}

Proposition~\ref{prop:block-coupling} gives the intended status of the
blockwise relaxation. When inter-block coupling is weak in the normalized sense
of \eqref{eq:block-coupling-coefficient}, blockwise curvature information is
close to the full-$\rmH$ oracle up to the displayed multiplicative factors.
When the coupling is large, the blockwise wrapper is not invalid; it gives up
coordinated use of off-diagonal curvature in exchange for a cheaper and more
stable randomization rule. Finally, the standard norm-matched ZO
forgetting-gap identity can still be read inside each block as a local shaping
score,
\begin{equation}
  \mathcal{S}_b^{\text{\ours}}
  \triangleq
  \frac{\eta^2}{2}
  \tau_b\norm{\vg_b}^{2}\left(\lamdir_b-\lamavg_b\right),
  \quad
  \lamdir_b\triangleq\frac{\vg_b^\top\rmH_{bb}\vg_b}{\norm{\vg_b}^{2}},
  \quad
  \lamavg_b\triangleq\frac{\tr(\rmH_{bb})}{d_b},
  \quad
  \tau_b\triangleq\frac{d_b}{q_b+d_b+1}.
  \label{eq:block-shaping-score}
\end{equation}
This score is not a pointwise decomposition of the full quadratic form. It is a
block-local prediction of where the \ours{} wrapper has positive within-block
anisotropic excess to attenuate after the mean step and scale have been
controlled. These quantities have distinct roles: $\mathcal{S}_b^{\text{\ours}}$
tracks the within-block anisotropic benefit predicted by
Theorem~\ref{thm:rise-blockwise-gap}; $D_b$ tracks the average cost of
isotropic randomized exposure; $C_b$ tracks sensitivity to omitted cross-block
coupling and one-batch instability; and $R_b(\rho)$ is an oracle-style
flat-signal proxy for whether the incoming block signal lies in low-retention
directions. None of these quantities is an unconditional selection threshold
on its own. Their selection value is assessed through the completed
query-count sweep and by additional scale controls when those controls are
reported, as discussed in Appendix~\ref{app:rise-evaluation-details}.

\section{Extended Experimental Details}
\label{app:extended-experiments}

The main experiments report compact transfer results, so the appendix separates
the evidence roles. The quadratic-sandbox study measures theorem-level
quantities directly; the end-to-end sections specify the visual and language
continual-learning settings; and unreported controls are described as supporting
analyses rather than evidence.

\subsection{Evidence Roles, Controls, and Settings}
\label{app:exp-setup}

Theory-facing experiments report operator residuals for
$\E\left[\rmP^\top\rmH\rmP\right]$ against the prediction in
\eqref{eq:anisotropy-attenuation}, the empirical damage gap
$\mathcal{Q}^{\text{FO}}-\E\left[\mathcal{Q}_q^{\text{ZO}}\right]$ against
\eqref{eq:direct-damage-gap}, centered finite-query deviations
$\mathcal{Q}_q^{\text{ZO}}-\E\left[\mathcal{Q}_q^{\text{ZO}}\right]$, and the
correlation between blockwise scores and observed \ours{} benefit.
Algorithm-facing experiments report final average accuracy, average forgetting,
new-task plasticity, backward transfer when applicable, training-time overhead,
and additional memory overhead. This separation prevents final accuracy from
becoming a proxy for theorem validation.

Mechanism controls include FO, scaled FO with the same mean-step shrinkage
induced by norm matching, FO plus isotropic Gaussian noise, FO plus
covariance-matched additive noise, global exact-gradient shaping, and raw
low-query ZO. For blockwise \ours{}, the mean-shrinkage control is blockwise
rather than global: with
$a_b\triangleq\sqrt{q_b/(q_b+d_b+1)}$ and
$\rmA\triangleq\operatorname{blkdiag}(a_1\rmI_1,\ldots,a_B\rmI_B)$, the
mean-matched FO control uses $\rmA\vg$. The covariance-matched noise control
uses independent block noise with covariance
\begin{equation}
  \rmC_b
  =
  \frac{1}{q_b+d_b+1}
  \left(\vg_b\vg_b^\top+\norm{\vg_b}^{2}\rmI_b\right),
  \label{eq:block-covariance-control}
\end{equation}
matching the centered covariance of $\rmP_b\vg_b$. These controls remove
smaller-step, blockwise shrinkage, generic-noise, and gradient-estimation
explanations before attributing a gain to randomized gradient shaping. Method
comparisons use a compact set of related continual learning baselines: GPM as
the projection-based geometry comparator \citep{saha2021gradient}, experience
replay as a strong simple memory baseline \citep{rebuffi2017icarl}, and a
ZO-FC-style hybrid comparator in PEFT settings where the classifier or head is
naturally optimized by FO gradients \citep{yu2025more}. Because \ours{} is a
post-processing rule on exact gradients, it can be attached to any host method
whose adaptation exposes suitable gradient blocks. The reported vision
instantiation uses ZO-FC as that host method, so the primary empirical claim is
paired improvement over ZO-FC; other host-method gains are treated as transfer
tests only when the corresponding \ours{} variant is explicitly evaluated.

For visual continual learning, we use CIFAR100, ImageNet-R, and DomainNet.
Following the settings in prior work, CIFAR100 and ImageNet-R are evaluated with
5-class and 10-class increments, and DomainNet is evaluated with 69 classes per
task. The class or task order is fixed with seed 1993 for all methods and
datasets. All visual experiments use a ViT-B/16 backbone. ZO-FC+\ours{} keeps
the ZO-FC convention of using an FO classifier head for fast plasticity, but
replaces the representation-side finite-difference ZO adaptation by the
exact-gradient shaping rule in Eq.~\eqref{eq:rise-block}. For the shaped
representation-side component, we use learning rate $0.01$ and $q=4$. The
classifier head is optimized with FO SGD using learning rate $0.01$ and a
cosine decay schedule. In the ZO-FC host, the representation-side adapter uses
SPSA with ZO-SGD, constant learning rate $0.01$, perturbation magnitude
$\epsilon=0.001$, query budget $q=4$, and $\ell_2$ clipping threshold $1.0$ for
the estimated gradients; ZO-FC+\ours{} keeps the same host classifier training
and replaces this finite-difference estimator by exact-gradient norm-matched
shaping. The CIFAR100 budget trains the FO classifier head for 10 epochs and
the representation-side adapter for 20 epochs; the ImageNet-R budget trains the
FO classifier head for 20 epochs and the adapter for 40 epochs. Baselines that
combine an adapter with a cosine classifier use the matched host budgets of 10
epochs on CIFAR100 and 20 epochs on ImageNet-R, while the other displayed
baselines use their official implementations and recommended hyperparameters.
This design makes Table~\ref{tab:rise-fc-results} a paired ZO-FC
ablation rather than a claim that \ours{} uniformly dominates all displayed
continual-learning baselines.

For language continual learning, we use the task pool summarized in
Table~\ref{tab:olora-language-datasets}, following the O-LoRA training setup
\citep{wang2023orthogonal}. The pool combines standard CL benchmark tasks,
GLUE tasks, SuperGLUE-style tasks, and IMDB movie reviews. Our three task
streams are listed in Table~\ref{tab:olora-language-orders}. Orders 1 and 2 are
short standard CL streams over four classification tasks, while Order 3 is a
heterogeneous 15-task stream spanning sentiment analysis, topic classification,
natural language inference, paraphrase detection, question answering, and
word-sense disambiguation. The experiments use T5 models; all variants use
AdamW, all tasks are trained for one epoch, and reported results are means over
three random seeds. We otherwise keep the same task streams, training budget,
batching, optimizer settings, and regularization settings when comparing FO,
ZO, and \ours{} variants. This keeps
the language-model transfer test focused on whether exact-gradient randomized
shaping recovers plasticity relative to finite-difference ZO while retaining
forgetting benefits.

\begin{table}[t]
  \centering
  \scriptsize
  \setlength{\tabcolsep}{3pt}
  \caption{Language continual-learning task pool used by the O-LoRA-style
  setting. The table restates the dataset coverage used for our language
  experiments with compact task and domain descriptions.}
  \begin{tabular}{@{}p{0.13\linewidth}p{0.17\linewidth}p{0.22\linewidth}p{0.22\linewidth}p{0.09\linewidth}@{}}
    \toprule
    \textbf{Dataset} & \textbf{Source} & \textbf{Task type} & \textbf{Domain} & \textbf{Metric} \\
    \midrule
    Yelp & CL benchmark & Sentiment analysis & Yelp reviews & Accuracy \\
    Amazon & CL benchmark & Sentiment analysis & Amazon reviews & Accuracy \\
    DBpedia & CL benchmark & Topic classification & Wikipedia & Accuracy \\
    Yahoo & CL benchmark & Topic classification & Yahoo Q\&A & Accuracy \\
    AG News & CL benchmark & Topic classification & News & Accuracy \\
    MNLI & GLUE & Natural language inference & Mixed genres & Accuracy \\
    QQP & GLUE & Paraphrase detection & Quora questions & Accuracy \\
    RTE & GLUE & Natural language inference & News and Wikipedia & Accuracy \\
    SST-2 & GLUE & Sentiment analysis & Movie reviews & Accuracy \\
    WiC & SuperGLUE & Word-sense disambiguation & Lexical examples & Accuracy \\
    CB & SuperGLUE & Natural language inference & Mixed genres & Accuracy \\
    COPA & SuperGLUE & Causal QA & Blogs and encyclopedia & Accuracy \\
    BoolQA & SuperGLUE & Boolean QA & Wikipedia & Accuracy \\
    MultiRC & SuperGLUE & Multi-sentence QA & Mixed passages & Accuracy \\
    IMDB & Movie-review dataset & Sentiment analysis & Movie reviews & Accuracy \\
    \bottomrule
  \end{tabular}
  \label{tab:olora-language-datasets}
\end{table}

\begin{table}[t]
  \centering
  \small
  \setlength{\tabcolsep}{4pt}
  \caption{Language task streams used in our experiments. The third stream is
  the long heterogeneous order used for the large-task-sequence setting.}
  \begin{tabular}{@{}p{0.16\linewidth}p{0.76\linewidth}@{}}
    \toprule
    \textbf{Stream} & \textbf{Task sequence} \\
    \midrule
    Order 1
    & DBpedia \(\rightarrow\) Amazon \(\rightarrow\) Yahoo \(\rightarrow\) AG News \\
    Order 2
    & DBpedia \(\rightarrow\) Amazon \(\rightarrow\) AG News \(\rightarrow\) Yahoo \\
    Order 3
    & Yelp \(\rightarrow\) Amazon \(\rightarrow\) MNLI \(\rightarrow\) CB
      \(\rightarrow\) COPA \(\rightarrow\) QQP \(\rightarrow\) RTE
      \(\rightarrow\) IMDB \(\rightarrow\) SST-2 \(\rightarrow\) DBpedia
      \(\rightarrow\) AG News \(\rightarrow\) Yahoo \(\rightarrow\) MultiRC
      \(\rightarrow\) BoolQA \(\rightarrow\) WiC \\
    \bottomrule
  \end{tabular}
  \label{tab:olora-language-orders}
\end{table}

Across all reported experiments, we use a single NVIDIA A800 GPU with 80 GB
memory and keep the training batch size fixed at 48.

\subsection{Mechanism Evidence in a Quadratic Sandbox}
\label{app:controlled-validation}

The quadratic-sandbox study uses a locally quadratic setting in which $\rmH$,
$\vg$, $q$, and $d$ are directly specified. For a fixed spectrum and query count,
we sample norm-matched ZO shapes, estimate
$\E\left[\rmP^\top\rmH\rmP\right]$, and compare the empirical average with the
closed form in \eqref{eq:anisotropy-attenuation}. The operator panel reports
eigenvalue-level contraction toward $\lamavg$ together with a Frobenius
operator residual as the number of Monte Carlo samples increases. The main
synthetic figure uses $d=64$, $q=4$, and $24{,}000$ Monte Carlo samples for the
eigenvalue panel; the relative eigenvalue error is $1.17\times 10^{-2}$.
Repeating the same validation after a random orthogonal rotation of $\rmH$ gives a
non-diagonal robustness test: Theorem~\ref{thm:anisotropy} is a distributional
isotropy statement, not a coordinate artifact.

The direct damage panel rotates $\vg$ relative to the eigenvectors of $\rmH$ so
that $\delta_\lambda=\lamdir-\lamavg$ sweeps through negative, near-zero, and
positive regimes. The plotted scalar is
$\mathcal{Q}^{\text{FO}}-\E\left[\mathcal{Q}_q^{\text{ZO}}\right]$, with the
predicted line from \eqref{eq:direct-damage-gap} overlaid. Negative and
near-zero regimes are part of the validation, because they verify that the
mechanism is geometry-dependent rather than a uniform protection guarantee. In
the main synthetic run, the empirical-versus-predicted damage gaps have
$R^2=0.9999$.

Finite-query validation compares the deterministic mean signal $G_q$ in
\eqref{eq:mean-gap-signal} together with the realized centered deviation across
query batches, dimensions, and block sizes. Increasing $q$
reduces realized variance, while large global dimensions make one-shot shaping
unstable; replacing the global shape by smaller blockwise shapes lowers the
realization scale. The main variance panel uses $d=256$ and compares global
shaping with blockwise shaping at $d_b=16$; at $q=2$, the blockwise standard
deviation is $0.79$ times the global standard deviation. This variance pattern
is the controlled transition from the global ZO identity to the blockwise
\ours{} wrapper.

\begin{figure}[t]
  \centering
  \includegraphics[width=\linewidth]{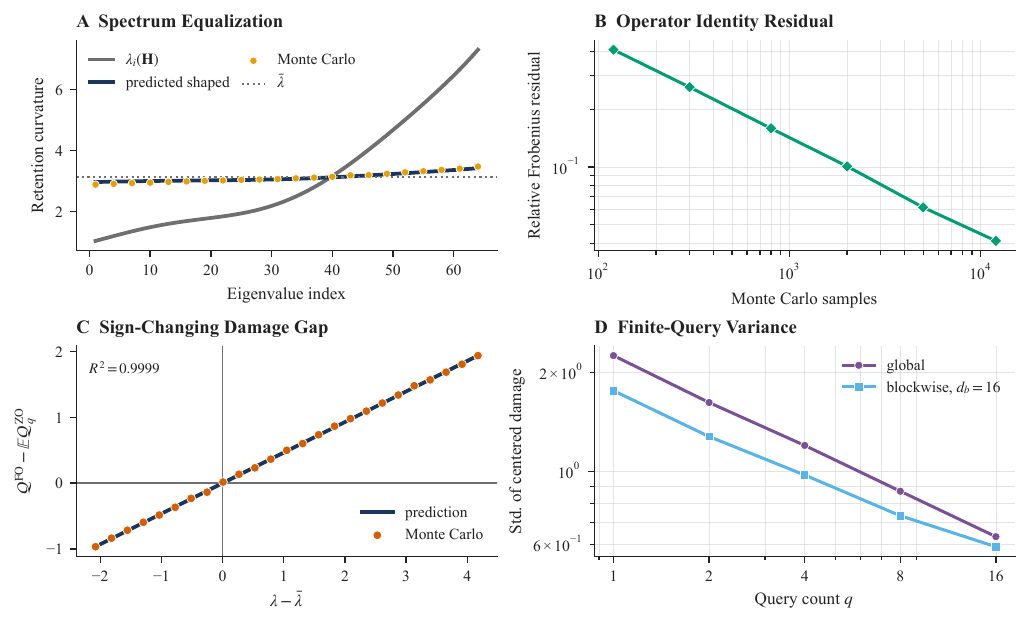}
  \caption{
  \textbf{Synthetic validation of randomized shaping.}
  In a controlled quadratic sandbox, norm-matched ZO shaping contracts the
  retention-curvature spectrum toward $\lamavg$ while preserving the mean
  curvature (A), and the Monte Carlo operator residual decreases with sample
  size (B). Rotating $\vg$ across eigendirections makes the direct damage gap
  change sign exactly as predicted by \eqref{eq:direct-damage-gap} (C, with
  $R^2=0.9999$ between empirical and predicted gaps). In a larger synthetic
  setting, blockwise shaping with $d_b=16$ lowers the centered one-batch damage
  deviation relative to global shaping across query counts (D, where at $q=2$
  the standard deviation is $0.79\times$ the global value).
  }
  \label{fig:synthetic-shaping}
\end{figure}

An optional neural local validation connects the sandbox to ordinary training. At
selected continual-learning checkpoints, historical batches define an empirical
Fisher or Hessian-style retention curvature proxy, the current batch supplies
the incoming exact gradient, and one-step or few-step measurements estimate the
resulting change in historical loss. The comparison is local: predicted
quadratic damage gaps and blockwise scores are correlated with
observed local forgetting changes. This local validation tests whether the mechanism
remains visible in neural-network neighborhoods without claiming a global
nonconvex training theorem.

When blockwise scores are reported, the proxy construction is fixed across
methods. Diagonal and block-diagonal readings may use empirical Fisher or
Gauss--Newton estimates from the same historical mini-batches; directional
quantities such as $\vg_b^\top\rmH_{bb}\vg_b$ may be estimated by Hessian-vector
or Fisher-vector products; and damped inverses in $R_b(\rho)$ use the same
$\rho$ across layers and methods. Off-diagonal coupling scores are reported
only when the corresponding block-pair products are estimated explicitly; when
they are omitted, the interpretation is restricted to within-block exposure.

\subsection{End-to-End Settings and Ablations}
\label{app:rise-evaluation-details}

The end-to-end experiments treat \ours{} as an exact-gradient transfer of the ZO
shape, not as a generic ZO method, replay method, or memory-saving
finite-difference optimizer. They are benchmark-transfer tests that complement
the quadratic-sandbox evidence in Fig.~\ref{fig:synthetic-shaping}. The reported
vision table inserts \ours{} into ZO-FC and therefore tests the paired
ZO-FC to ZO-FC+\ours{} change. The language table uses SeqLoRA and O-LoRA as
additional hosts, illustrating that the same wrapper can be attached beyond
ZO-FC when exact block gradients are available.

For visual continual learning, the central comparison is ZO-FC versus
ZO-FC+\ours{}. Both methods keep the classifier head optimized by FO gradients,
so the head remains a fast-plastic component. The difference is on the
representation side: ZO-FC uses finite-difference ZO adaptation, while
ZO-FC+\ours{} applies the exact-gradient shaping rule in
Eq.~\eqref{eq:rise-block}.
The classifier-only, FO-optimization, and ZO-optimization rows in
Table~\ref{tab:rise-fc-results} provide context for this paired comparison.
They are not used to claim generic dominance over all continual-learning
methods.

For language continual learning, Table~\ref{tab:additional-cl-benchmarks}
compares FO, finite-difference ZO, and \ours{} variants under the same task
orders and training settings for SeqLoRA and O-LoRA. This design tests the
specific confound raised by recent ZO continual-learning results: raw
finite-difference ZO can reduce forgetting while losing plasticity. The reported
comparison tests whether exact-gradient transfer recovers much of that
plasticity while retaining some of the stability benefit associated with ZO
shaping. These rows support the attachable-wrapper view, but the magnitude of
the gain remains host-method dependent and is not inferred for methods without a
reported \ours{} variant.

Across both domains, the reported metrics are Avg, Last, and Fgt. Avg measures
overall continual-learning performance, Last serves as the plasticity-sensitive
readout, and Fgt measures retention. Efficiency breakdowns and memory overhead
are useful system-level measurements, but they are not part of the reported
evidence package.

The completed end-to-end ablation is the query-count sweep in
Table~\ref{tab:q-ablation-imagenet-r}. It varies
$q\in\{1,4,8,16,32,64\}$ on ImageNet-R Inc-5 while keeping the rest of the
visual setting fixed. This ablation tests the finite-query prediction that low
$q$ gives stronger randomized shaping and lower forgetting, while larger $q$
moves the adaptation closer to the exact-gradient direction and improves
plasticity-sensitive accuracy.

\begin{table}[t]
  \centering
  \setlength{\tabcolsep}{14pt}
  \caption{Ablation over query count $q$ on ImageNet-R Inc-5.}
  \begin{tabular}{@{}rrrr@{}}
    \toprule
    $q$ & Avg$\uparrow$ & Last$\uparrow$ & Fgt$\downarrow$ \\
    \midrule
    1 & 71.21 & 66.42 & \textbf{2.63} \\
    4 & 72.05 & 67.57 & \underline{2.68} \\
    8 & 72.98 & 68.27 & 3.48 \\
    16 & 73.92 & 68.73 & 4.29 \\
    32 & \underline{74.94} & \textbf{69.27} & 5.08 \\
    64 & \textbf{75.06} & \underline{69.03} & 6.46 \\
    \bottomrule
  \end{tabular}
  \label{tab:q-ablation-imagenet-r}
\end{table}

Additional controls remain important for stronger claims, but they are
supporting analyses unless their results are reported. These include scaled
FO, FO plus isotropic Gaussian noise, FO plus covariance-matched noise, global
versus blockwise exact-gradient shaping, block-granularity sweeps,
parameter-scope ablations, and score-correlation analyses using the
blockwise scores in Appendix~\ref{app:rise-block-theory}. When such controls are
reported, they use the same task streams, seeds, adaptation budgets, and
optimizer placement as the corresponding end-to-end table. Their role is to
separate randomized-shaping benefits from mean shrinkage, generic noise,
implementation granularity, and score-selection effects.

The blind-optimal exposure result in Sec.~\ref{sec:theory-scope} remains a
supporting scope analysis. A compact exposure panel comparing FO, norm-matched
ZO, and the isotropic second-moment benchmark can clarify the unknown-curvature
interpretation, but it does not replace the fixed-$\rmH$ operator identity,
direct forgetting-gap validation, or the reported end-to-end transfer results.

\section{Proofs}
\label{app:proofs}

The proofs follow the dependency order used by the main text: finite-difference
calibration, norm matching, expected shaped curvature, direct damage,
one-batch corrections, curvature-agnostic exposure, and blockwise \ours{}
support.

\subsection{Proof of Lemma~\ref{lem:finite-smoothing}}

By $L$-smoothness, for each queried direction $\rvz_r$ there exist scalar
remainders $\epsilon_{r,+}$ and $\epsilon_{r,-}$ such that
\begin{align}
  f(\vtheta+\mu\rvz_r)
  &=
  f(\vtheta)
  +\mu\langle\vg,\rvz_r\rangle
  +\epsilon_{r,+},
  &
  |\epsilon_{r,+}|
  &\le
  \frac{L\mu^2}{2}\norm{\rvz_r}^{2},\\
  f(\vtheta-\mu\rvz_r)
  &=
  f(\vtheta)
  -\mu\langle\vg,\rvz_r\rangle
  +\epsilon_{r,-},
  &
  |\epsilon_{r,-}|
  &\le
  \frac{L\mu^2}{2}\norm{\rvz_r}^{2}.
\end{align}
Thus
\begin{equation}
  \frac{f(\vtheta+\mu\rvz_r)-f(\vtheta-\mu\rvz_r)}{2\mu}
  =
  \langle\vg,\rvz_r\rangle+\epsilon_r,
  \qquad
  \epsilon_r
  \triangleq
  \frac{\epsilon_{r,+}-\epsilon_{r,-}}{2\mu},
\end{equation}
with
\begin{equation}
  |\epsilon_r|
  \le
  \frac{L\mu}{2}\norm{\rvz_r}^{2}.
\end{equation}
Substituting the two-point Taylor decomposition into
\eqref{eq:two-point-estimator} yields
\begin{equation}
  \widehat{\vg}_{\mu}
  =
  \frac{1}{q}\sum_{r=1}^{q}
  \rvz_r\rvz_r^\top\vg
  +
  \frac{1}{q}\sum_{r=1}^{q}\epsilon_r\rvz_r
  =
  \rmZ\vg+\vr_{\mu},
  \qquad
  \vr_{\mu}\triangleq\frac{1}{q}\sum_{r=1}^{q}\epsilon_r\rvz_r.
\end{equation}
Conditioned on the queried directions,
\begin{equation}
  \norm{\vr_{\mu}}
  \le
  \frac{1}{q}\sum_{r=1}^{q}|\epsilon_r|\,\norm{\rvz_r}
  \le
  \frac{L\mu}{2q}\sum_{r=1}^{q}\norm{\rvz_r}^{3}.
\end{equation}
Taking expectation gives
\begin{equation}
  \E\left[\norm{\vr_{\mu}}\right]
  \le
  \frac{L\mu}{2}\,
  \E_{\rvz\sim\mathcal{N}(\vzero,\rmI)}\left[\norm{\rvz}^{3}\right].
\end{equation}
The Gaussian fourth moment gives the simple dimension bound
\begin{equation}
  \E\left[\norm{\rvz}^{3}\right]
  \le
  \left(\E\left[\norm{\rvz}^{4}\right]\right)^{3/4}
  =
  \left(d(d+2)\right)^{3/4}
  =
  \mathcal{O}\left(d^{3/2}\right).
\end{equation}
For the second moment, Jensen's inequality gives
$q^{-1}\sum_{r=1}^q\norm{\rvz_r}^{3}$ squared no larger than
$q^{-1}\sum_{r=1}^q\norm{\rvz_r}^{6}$. Hence the Gaussian sixth moment gives
\begin{equation}
  \E\left[\norm{\vr_{\mu}}^{2}\right]
  \le
  \frac{L^2\mu^2}{4}\,
  \E_{\rvz\sim\mathcal{N}(\vzero,\rmI)}\left[\norm{\rvz}^{6}\right]
  =
  \mathcal{O}\left(L^2\mu^2 d^3\right).
\end{equation}
Together these bounds prove \eqref{eq:finite-mu-bridge}.

\subsection{Proof of Lemma~\ref{lem:mean-norm}}

Since $\E\left[\rvz_r\rvz_r^\top\right]=\rmI$, linearity gives
\begin{equation}
  \E\left[\rmZ\right]
  =
  \frac{1}{q}\sum_{r=1}^q
  \E\left[\rvz_r\rvz_r^\top\right]
  =
  \rmI.
\end{equation}
For the second moment, expand
\begin{equation}
  \E\left[\norm{\rmZ\vg}^2\right]
  =
  \frac{1}{q^2}
  \sum_{r,s=1}^q
  \E\left[\vg^\top \rvz_r \rvz_r^\top \rvz_s \rvz_s^\top \vg\right].
\end{equation}
For $r\neq s$, independence gives
\begin{equation}
  \E\left[\vg^\top \rvz_r \rvz_r^\top \rvz_s \rvz_s^\top \vg\right]
  =
  \vg^\top
  \E\left[\rvz_r\rvz_r^\top\right]
  \E\left[\rvz_s\rvz_s^\top\right]\vg
  =
  \norm{\vg}^2.
\end{equation}
For $r=s$, rotational invariance and the Gaussian fourth moment give
\begin{equation}
  \E\left[\norm{\rvz_r}^2(\rvz_r^\top \vg)^2\right]=(d+2)\norm{\vg}^2.
\end{equation}
Combining the $q$ diagonal and $q(q-1)$ off-diagonal terms yields
\begin{equation}
  \E\left[\norm{\rmZ\vg}^2\right]
  =
  \frac{q(d+2)+q(q-1)}{q^2}\norm{\vg}^2
  =
  \frac{q+d+1}{q}\norm{\vg}^2.
\end{equation}
Multiplying by $\kappa^{-1}=q/(q+d+1)$ gives
$\E[\norm{\rmP\vg}^{2}]=\norm{\vg}^{2}$, and linearity gives
$\E[\rmP\vg]=\kappa^{-1/2}\vg=\sqrt{q/(q+d+1)}\,\vg$. This proves
\eqref{eq:norm-matched-calibration}.

\subsection{Full-Information Isotropic Reference}

Let $\lamavg=\tr(\rmH)/d$ and
$\rmP_\star=\sqrt{\tr(\rmH)/d}\,\rmH^{-1/2}\rmO$ with $\rmO^\top\rmO=\rmI$.
Since $\rmH\succ0$, this operator is well defined. Direct substitution gives
\begin{equation}
  \rmP_\star^\top\rmH\rmP_\star
  =
  \frac{\tr(\rmH)}{d}\rmO^\top\rmH^{-1/2}\rmH\rmH^{-1/2}\rmO
  =
  \lamavg\rmI.
\end{equation}
Taking traces yields
$\tr(\rmP_\star^\top\rmH\rmP_\star)=\tr(\lamavg\rmI)=\tr(\rmH)$, which proves
both the fixed-scale identity and the isotropic target. For any
$\vg\neq\vzero$,
\begin{equation}
  \mathcal{Q}(-\eta\rmP_\star\vg)
  =
  \frac{\eta^2}{2}\vg^\top
  \left(\rmP_\star^\top\rmH\rmP_\star\right)\vg
  =
  \frac{\eta^2}{2}\lamavg\norm{\vg}^{2}.
\end{equation}
Subtracting this expression from
$\mathcal{Q}^{\text{FO}}=(\eta^2/2)\norm{\vg}^{2}\lamdir$ gives
\[
  \mathcal{Q}^{\text{FO}}
  -
  \mathcal{Q}(-\eta\rmP_\star\vg)
  =
  \frac{\eta^2}{2}\norm{\vg}^{2}\left(\lamdir-\lamavg\right),
\]
which is the full-information damage gap stated informally in
Section~\ref{sec:anisotropy}, with
$\lamdir=\vg^\top\rmH\vg/\norm{\vg}^{2}$.

\subsection{A General Isotropic-Moment Lemma}

\begin{lemma}[General Isotropic-Moment Form]
\label{lem:app-isotropic-moment-form}
Let $\rmP$ be a random shape operator with finite second moments whose distribution is
isotropic in a selected comparison space, meaning
$\rmO^\top\rmP\rmO\stackrel{d}{=}\rmP$ for every orthogonal matrix $\rmO$,
where $\stackrel{d}{=}$ denotes equality in distribution. Define
$\gT(\rmH)\triangleq\E\left[\rmP^\top\rmH\rmP\right]$ and
$\lamavg=\tr(\rmH)/d$. Then there exist scalars $\alpha$ and $\rho$, depending
on the shape-operator distribution but not on $\rmH$, such that for every
symmetric $\rmH$,
\begin{equation}
  \gT(\rmH)
  =
  \alpha\rmH+\rho\lamavg\rmI.
  \label{eq:universality}
\end{equation}
Equivalently, if $\rmH=\lamavg\rmI+\rmR$ with $\tr(\rmR)=0$, where $\rmR$ is
the traceless anisotropic component of retention curvature, then
\begin{equation}
  \gT(\rmH)
  =
  (\alpha+\rho)\lamavg\rmI+\alpha\rmR.
  \label{eq:universality-traceless}
\end{equation}
\end{lemma}

The isotropic-moment assumption in Lemma~\ref{lem:app-isotropic-moment-form} implies that
the linear map $\gT(\rmH)=\E\left[\rmP^\top\rmH\rmP\right]$ is orthogonally equivariant:
$\gT(\rmO^\top\rmH\rmO)=\rmO^\top\gT(\rmH)\rmO$ for every orthogonal $\rmO$.
The space of symmetric matrices decomposes into the direct sum of the identity
span and the traceless symmetric subspace. These two subspaces are invariant
under orthogonal conjugation. By the standard irreducibility of the traceless
symmetric representation, an orthogonally equivariant linear map acts as a
scalar on each component. Hence there exist scalars $\omega$ and $\alpha$ such that
\begin{equation}
  \gT(\lamavg\rmI+\rmR)
  =
  \omega\lamavg\rmI+\alpha\rmR,
  \qquad \tr(\rmR)=0.
\end{equation}
Writing $\rho=\omega-\alpha$ gives
$\gT(\rmH)=\alpha\rmH+\rho\lamavg\rmI$, which is
\eqref{eq:universality}; the equivalent traceless form follows immediately.

\subsection{Proof of Theorem~\ref{thm:anisotropy}}

We expand
\begin{equation}
  \E\left[\rmZ\rmH\rmZ\right]
  =
  \frac{1}{q^2}
  \sum_{r,s=1}^q
  \E\left[\rvz_r\rvz_r^\top \rmH \rvz_s\rvz_s^\top\right].
\end{equation}
For $r\neq s$, independence gives $\rmH$. For $r=s$, the Gaussian fourth-moment
identity gives
\begin{equation}
  \E\left[\rvz\rvz^\top \rmH\rvz\rvz^\top\right]=\tr(\rmH)\rmI+2\rmH.
\end{equation}
Therefore,
\begin{equation}
  \E\left[\rmZ\rmH\rmZ\right]
  =
  \frac{1}{q^2}
  \left(q(\tr(\rmH)\rmI+2\rmH)+q(q-1)\rmH\right)
  =
  \frac{q+1}{q}\rmH+\frac{\tr(\rmH)}{q}\rmI.
\end{equation}
Since $\rmZ$ and $\rmP=\kappa^{-1/2}\rmZ$ are symmetric,
\begin{equation}
  \E\left[\rmP^\top\rmH\rmP\right]
  =
  \frac{q}{q+d+1}
  \left(
    \frac{q+1}{q}\rmH+\frac{\tr(\rmH)}{q}\rmI
  \right)
  =
  \frac{q+1}{q+d+1}\rmH
  +
  \frac{\tr(\rmH)}{q+d+1}\rmI.
\end{equation}
Using $\lamavg=\tr(\rmH)/d$ and $\tau=d/(q+d+1)$ gives
\begin{equation}
  \E\left[\rmP^\top\rmH\rmP\right]
  =
  (1-\tau)\rmH+\tau\lamavg \rmI.
\end{equation}

\subsection{Proof of Corollary~\ref{cor:spectral}}

Since $\rmI$ commutes with $\rmH$, the matrix
$\E\left[\rmP^\top\rmH\rmP\right]=(1-\tau)\rmH+\tau\lamavg\rmI$ has the same
eigenvectors as $\rmH$, and its eigenvalues are exactly
\begin{equation}
  \lambda'_i=(1-\tau)\lambda_i+\tau\lamavg,
  \qquad i=1,\ldots,d.
\end{equation}
The map $\vlambda\mapsto(1-\tau)\vlambda+\tau\lamavg\vone$ preserves the sum of
eigenvalues and contracts every deviation from the mean by $1-\tau\in(0,1)$.

\subsection{Proof of Theorem~\ref{thm:direct-damage}}

By the definition of $\mathcal{Q}_q^{\text{ZO}}$ and
Theorem~\ref{thm:anisotropy},
\begin{equation}
  \E\left[\mathcal{Q}_q^{\text{ZO}}\right]
  =
  \frac{\eta^2}{2}
  \vg^\top
  \left((1-\tau)\rmH+\tau\lamavg\rmI\right)
  \vg .
\end{equation}
Using
$\vg^\top\rmH\vg=\norm{\vg}^{2}\lamdir$ and
$\vg^\top\rmI\vg=\norm{\vg}^{2}$ gives
\begin{equation}
  \E\left[\mathcal{Q}_q^{\text{ZO}}\right]
  =
  \frac{\eta^2}{2}\norm{\vg}^{2}
  \left((1-\tau)\lamdir+\tau\lamavg\right).
\end{equation}
Since
$\mathcal{Q}^{\text{FO}}=(\eta^2/2)\norm{\vg}^{2}\lamdir$, subtracting the last display
from $\mathcal{Q}^{\text{FO}}$ gives
\begin{equation}
  \mathcal{Q}^{\text{FO}}-\E\left[\mathcal{Q}_q^{\text{ZO}}\right]
  =
  \frac{\eta^2}{2}\tau\norm{\vg}^{2}
  \left(\lamdir-\lamavg\right),
\end{equation}
which is the claimed forgetting-reduction identity. If $\lamdir>0$, then
$\mathcal{Q}^{\text{FO}}=(\eta^2/2)\norm{\vg}^{2}\lamdir>0$, and dividing the
last display by $\mathcal{Q}^{\text{FO}}$ gives
$\tau(1-\lamavg/\lamdir)$.

\subsection{Proof of Proposition~\ref{prop:finite-query-deviation}}

Let $\rmD\triangleq\rmZ-\rmI$ and
$\Xi_q\triangleq\rmP^\top\rmH\rmP-\bar{\rmH}_{q}$. Also let
$\lambda_{\max}$ denote the largest eigenvalue of $\rmH$. Since
$\rmP=\kappa^{-1/2}\rmZ$, $\rmZ=\rmI+\rmD$, and $\rmH$ is positive
semidefinite,
\begin{equation}
  \rmP^\top\rmH\rmP
  =
  \kappa^{-1}\rmZ\rmH\rmZ
  =
  \kappa^{-1}
  \left(\rmH+\rmD\rmH+\rmH\rmD+\rmD\rmH\rmD\right).
  \label{eq:app-realized-operator}
\end{equation}
The proof of Theorem~\ref{thm:anisotropy} gives
\begin{equation}
  \bar{\rmH}_{q}
  =
  \E\left[\rmP^\top\rmH\rmP\right]
  =
  \kappa^{-1}
	  \left[
	  \rmH+\frac{1}{q}\left(\rmH+d\lamavg\rmI\right)
	  \right].
  \label{eq:app-mean-operator}
\end{equation}
Subtracting \eqref{eq:app-mean-operator} from
\eqref{eq:app-realized-operator} gives the exact decomposition
\begin{equation}
  \Xi_q
  =
  \kappa^{-1}
  \left[
	  \rmD\rmH+\rmH\rmD+\rmD\rmH\rmD
	  -
	  \frac{1}{q}\left(\rmH+d\lamavg\rmI\right)
	  \right].
  \label{eq:finite-query-operator-decomp}
\end{equation}
The zero-smoothing realized damage identity used in
Proposition~\ref{prop:finite-query-deviation} follows from
\begin{equation}
  \mathcal{Q}_q^{\text{ZO}}
  =
  \frac{\eta^2}{2}\vg^\top
  \left(\bar{\rmH}_{q}+\Xi_q\right)\vg
  =
  \E\left[\mathcal{Q}_q^{\text{ZO}}\right]
  +
  \frac{\eta^2}{2}\vg^\top\Xi_q\vg
\end{equation}
and Theorem~\ref{thm:direct-damage}. Equivalently,
\begin{equation}
  \left|
  \left(\mathcal{Q}^{\text{FO}}-\mathcal{Q}_q^{\text{ZO}}\right)
  -
  G_q
  \right|
  =
  \left|
  \mathcal{Q}_q^{\text{ZO}}
  -
  \E\left[\mathcal{Q}_q^{\text{ZO}}\right]
  \right|
  =
  \frac{\eta^2}{2}\left|\vg^\top\Xi_q\vg\right|.
\end{equation}

It remains to prove the high-probability operator-norm bound. A standard
Gaussian sample-covariance concentration inequality gives a universal constant
$C>0$ such that, with probability at least $1-\delta$,
\begin{equation}
  \norm{\rmD}_{\text{op}}
  \le
  \epsilon_q,
  \qquad
  \epsilon_q
  =
  C\left[
  \sqrt{\frac{d+\log(1/\delta)}{q}}
  +
  \frac{d+\log(1/\delta)}{q}
  \right].
  \label{eq:app-operator-covariance}
\end{equation}
Increasing $C$ if necessary, $\epsilon_q\le C\psi_q(\delta)$ for
$\psi_q(\delta)$ in \eqref{eq:finite-query-fluctuation-scale}. On the same
event, the random part in \eqref{eq:finite-query-operator-decomp} satisfies
\begin{equation}
  \norm{\rmD\rmH+\rmH\rmD+\rmD\rmH\rmD}_{\text{op}}
  \le
  \left(2\epsilon_q+\epsilon_q^2\right)\lambda_{\max}.
\end{equation}
The deterministic centering term is absorbed by the same fluctuation scale:
since $\rmH\succeq0$,
$d\lamavg=\tr(\rmH)\le d\lambda_{\max}$, and therefore
\begin{equation}
  \left\|
  \frac{1}{q}\left(\rmH+d\lamavg\rmI\right)
  \right\|_{\text{op}}
  \le
  \frac{\lambda_{\max}+d\lamavg}{q}
  \le
  \frac{d+1}{q}\lambda_{\max}
  \le
  \lambda_{\max}\psi_q(\delta).
\end{equation}
Combining these estimates with $\kappa^{-1}\le1$ gives
\begin{align}
  \norm{\Xi_q}_{\text{op}}
  &\le
  \kappa^{-1}
  \left(
    \norm{\rmD\rmH+\rmH\rmD+\rmD\rmH\rmD}_{\text{op}}
    +
    \frac{\lambda_{\max}+d\lamavg}{q}
  \right) \\
  &\le
  \kappa^{-1}
  \left[
    \left(2\epsilon_q+\epsilon_q^2\right)\lambda_{\max}
    +
    \frac{\lambda_{\max}+d\lamavg}{q}
  \right] \\
  &\le
  C\lambda_{\max}
  \left(\psi_q(\delta)+\psi_q(\delta)^2\right).
\end{align}
Using
$|\vg^\top\Xi_q\vg|\le\norm{\vg}^{2}\norm{\Xi_q}_{\text{op}}$ gives the
main-text bound in \eqref{eq:finite-query-scalar-bound}.

\subsection{Proof of Corollary~\ref{cor:finite-smoothing-damage}}

Let
$\widehat{\rvx}_{\mu}=\kappa^{-1/2}\widehat{\vg}_{\mu}$,
$\rvx=\rmP\vg=\kappa^{-1/2}\rmZ\vg$, and
$\vb=\kappa^{-1/2}\vr_{\mu}$. Lemma~\ref{lem:finite-smoothing} gives
$\widehat{\rvx}_{\mu}=\rvx+\vb$. By definition,
$\mathcal{Q}_q^{\text{ZO}}=\mathcal{Q}(-\eta\rvx)$ and
$\widehat{\mathcal{Q}}_{q,\mu}^{\text{ZO}}
=\mathcal{Q}(-\eta\widehat{\rvx}_{\mu})$. Adding and subtracting
$\mathcal{Q}_q^{\text{ZO}}$ in
$\mathcal{Q}^{\text{FO}}-\widehat{\mathcal{Q}}_{q,\mu}^{\text{ZO}}$ and then
using Theorem~\ref{thm:direct-damage} gives
\eqref{eq:practical-finite-mu-accounting}.

It remains to bound the final perturbation term. Since $\rmH\succeq0$ and
$\lambda_{\max}$ is its largest eigenvalue,
\begin{equation}
\begin{aligned}
  \left|
  \mathcal{Q}(-\eta\widehat{\rvx}_{\mu})
  -
  \mathcal{Q}(-\eta\rvx)
  \right|
  &=
  \frac{\eta^2}{2}
  \left|2\rvx^\top\rmH\vb+\vb^\top\rmH\vb\right| \\
  &\le
  \frac{\eta^2}{2}\lambda_{\max}
  \left(2\norm{\rvx}\norm{\vb}+\norm{\vb}^{2}\right).
\end{aligned}
\end{equation}
This pathwise inequality is the source of the expectation bound after
substituting $\rvx=\rmP\vg$ and $\vb=\kappa^{-1/2}\vr_{\mu}$.

Taking expectations and applying Cauchy--Schwarz,
\begin{equation}
  \E\left[\norm{\rmP\vg}\norm{\vr_{\mu}}\right]
  \le
  \left(\E\left[\norm{\rmP\vg}^{2}\right]\right)^{1/2}
  \left(\E\left[\norm{\vr_{\mu}}^{2}\right]\right)^{1/2}.
\end{equation}
Lemma~\ref{lem:mean-norm} gives
$\E[\norm{\rmP\vg}^{2}]=\norm{\vg}^{2}$, and
Lemma~\ref{lem:finite-smoothing} gives the required second-moment remainder
bound. The Cauchy--Schwarz estimate controls the cross term, while the residual
quadratic term contributes $\kappa^{-1}\E[\norm{\vr_{\mu}}^{2}]$ directly.
Substituting these estimates gives
$\E\left[|\widehat{\mathcal{Q}}_{q,\mu}^{\text{ZO}}
-\mathcal{Q}_q^{\text{ZO}}|\right]$ at the displayed order, proving
\eqref{eq:finite-smoothing-damage}.

For the squared-norm perturbation, use
$\widehat{\rvx}_{\mu}=\rvx+\vb$ and
$\E[\norm{\rvx}^{2}]=\norm{\vg}^{2}$:
\begin{equation}
\begin{aligned}
  \left|
  \E\left[\norm{\widehat{\rvx}_{\mu}}^{2}\right]
  -
  \norm{\vg}^{2}
  \right|
  &=
  \left|
  \E\left[2\rvx^\top\vb+\norm{\vb}^{2}\right]
  \right| \\
  &\le
  2\E[\norm{\rvx}\norm{\vb}]
  +
  \E[\norm{\vb}^{2}].
\end{aligned}
\end{equation}
The same Cauchy--Schwarz step and the same bounds on
$\vr_\mu$ give \eqref{eq:finite-smoothing-norm-mismatch}.

\subsection{Proof of Corollary~\ref{cor:realized-gap-sign}}

Proposition~\ref{prop:finite-query-deviation} gives, with probability at least
$1-\delta_q$, the zero-smoothing centered-deviation bound
$|\mathcal{Q}_q^{\text{ZO}}-\E[\mathcal{Q}_q^{\text{ZO}}]|
\le B_q(\delta_q)$. Corollary~\ref{cor:finite-smoothing-damage} gives
$\E[|\widehat{\mathcal{Q}}_{q,\mu}^{\text{ZO}}
-\mathcal{Q}_q^{\text{ZO}}|]\le B_\mu$, so Markov's inequality gives
\begin{equation}
  \left|
  \widehat{\mathcal{Q}}_{q,\mu}^{\text{ZO}}
  -
  \mathcal{Q}_q^{\text{ZO}}
  \right|
  \le
  \frac{B_\mu}{\delta_\mu}
\end{equation}
with probability at least $1-\delta_\mu$. Applying a union bound and the exact
accounting identity \eqref{eq:practical-finite-mu-accounting} proves
\eqref{eq:realized-gap-certificate}. If the right-hand side is smaller than
$|G_q|$, the realized gap cannot cross zero, so its sign agrees with $G_q$.

\subsection{Proof of Theorem~\ref{thm:hblind-isotropization}}

Fix $\rmM\succeq0$ and let $\lambda_{\max}(\rmM)$ be its largest eigenvalue.
Then $\rmM\preceq\lambda_{\max}(\rmM)\rmI$, so
$\lambda_{\max}(\rmM)\rmI-\rmM\succeq0$. For every
$\rmH\in\mathcal{H}_{\bar\lambda}$, the trace of the product of two positive
semidefinite matrices is nonnegative, because
$\tr(\rmH(\lambda_{\max}(\rmM)\rmI-\rmM))
=\tr(\rmH^{1/2}(\lambda_{\max}(\rmM)\rmI-\rmM)\rmH^{1/2})\ge0$. Hence
\begin{equation}
  \tr(\rmH \rmM)
  \le
  \lambda_{\max}(\rmM)\tr(\rmH)
  =
  d\bar\lambda\,\lambda_{\max}(\rmM).
\end{equation}
This upper bound is attained by choosing
$\rmH=d\bar\lambda\,\rvu\rvu^\top$, where $\rvu$ is a unit top eigenvector of
$\rmM$. Hence
\begin{equation}
  \sup_{\rmH\in\mathcal{H}_{\bar\lambda}}\tr(\rmH \rmM)
  =
  d\bar\lambda\,\lambda_{\max}(\rmM).
\end{equation}
Multiplying by $\eta^2/2$ proves \eqref{eq:hblind-worst-exposure}.
The minimax problem is therefore equivalent to minimizing $\lambda_{\max}(\rmM)$
over positive semidefinite $\rmM$ with $\tr(\rmM)=\norm{\vg}^{2}$. For any such $\rmM$,
\begin{equation}
  \lambda_{\max}(\rmM)
  \ge
  \frac{\tr(\rmM)}{d}
  =
  \frac{\norm{\vg}^{2}}{d}.
\end{equation}
Equality holds if and only if all eigenvalues of $\rmM$ are equal, which gives the
unique optimizer $\rmM^\star=(\norm{\vg}^{2}/d)\rmI$. Multiplying by $\eta^2/2$
proves the minimax equality in \eqref{eq:hblind-worst-exposure}.

\subsection{Appendix-Only Aligned Curvature-Agnostic Benchmark}
\label{app:aligned-hblind}

\begin{theorem}[Aligned-Signal Curvature-Agnostic Benchmark]
\label{thm:aligned-blind-optimum}
Assume $d\ge2$, fix $\alpha\in[0,1]$, and define
$\Pi_{\perp}\triangleq \rmI-\vg\vg^\top/\norm{\vg}^{2}$. Let
\begin{equation}
  \mathcal{M}_{\alpha}
  \triangleq
  \left\{
  \rmM\succeq0:
  \tr(\rmM)=\norm{\vg}^{2},\
  \rmM\succeq \alpha^{2}\vg\vg^\top
  \right\}.
  \label{eq:aligned-moment-set}
\end{equation}
For $\mathcal{H}_{\bar\lambda}$ in \eqref{eq:hblind-sets},
\begin{equation}
  V_{\alpha}^{\star}
  \triangleq
  \inf_{\rmM\in\mathcal{M}_{\alpha}}\ 
  \sup_{\rmH\in\mathcal{H}_{\bar\lambda}}
  \frac{\eta^2}{2}\tr(\rmH \rmM)
  =
  \frac{\eta^2}{2}d\bar\lambda\norm{\vg}^{2}
  \max\left\{\alpha^2,\frac{1}{d}\right\}.
  \label{eq:aligned-minimax-value}
\end{equation}
One optimal second moment is
\begin{equation}
  \rmM_{\alpha}^{\star}
  =
  \begin{cases}
  \displaystyle
  \frac{\norm{\vg}^{2}}{d}\rmI,
  & \alpha^2\le 1/d, \\[1.0ex]
  \displaystyle
  \alpha^2\vg\vg^\top
  +
  \frac{(1-\alpha^2)\norm{\vg}^{2}}{d-1}\Pi_{\perp},
  & \alpha^2> 1/d .
  \end{cases}
  \label{eq:aligned-minimax-moment}
\end{equation}
\end{theorem}

The set $\mathcal{M}_{\alpha}$ contains the second moments of all random
directions $\rvx$ satisfying $\E[\rvx]=\alpha\vg$ and
$\E\left[\norm{\rvx}^{2}\right]=\norm{\vg}^{2}$. The theorem records the aligned
curvature-agnostic characterization that the main text uses only as positioning.

\paragraph{Proof.}

The adversarial-curvature calculation from
Theorem~\ref{thm:hblind-isotropization} gives
\begin{equation}
  \sup_{\rmH\in\mathcal{H}_{\bar\lambda}}\tr(\rmH \rmM)
  =
  d\bar\lambda\,\lambda_{\max}(\rmM).
\end{equation}
It remains to minimize $\lambda_{\max}(\rmM)$ over
$\mathcal{M}_{\alpha}$. Every feasible $\rmM$ satisfies both
\begin{equation}
  \lambda_{\max}(\rmM)
  \ge
  \frac{\tr(\rmM)}{d}
  =
  \frac{\norm{\vg}^{2}}{d}
\end{equation}
and, by applying the constraint
$\rmM\succeq\alpha^2\vg\vg^\top$ to the unit vector
$\vg/\norm{\vg}$,
\begin{equation}
  \lambda_{\max}(\rmM)
  \ge
  \alpha^2\norm{\vg}^{2}.
\end{equation}
Thus
\begin{equation}
  \lambda_{\max}(\rmM)
  \ge
  \norm{\vg}^{2}\max\left\{\alpha^2,\frac{1}{d}\right\}.
\end{equation}
When $\alpha^2\le1/d$, the isotropic moment
$(\norm{\vg}^{2}/d)\rmI$ is feasible because
$(\norm{\vg}^{2}/d)\rmI-\alpha^2\vg\vg^\top\succeq0$, and it attains this lower
bound. When $\alpha^2>1/d$, the moment in
\eqref{eq:aligned-minimax-moment} has trace $\norm{\vg}^{2}$ and satisfies
$\rmM_{\alpha}^{\star}-\alpha^2\vg\vg^\top
=((1-\alpha^2)\norm{\vg}^{2}/(d-1))\Pi_\perp\succeq0$. Its eigenvalue along
$\vg$ is $\alpha^2\norm{\vg}^{2}$, and all perpendicular eigenvalues equal
$((1-\alpha^2)\norm{\vg}^{2})/(d-1)$, which is no larger because
$\alpha^2>1/d$. It therefore attains the lower bound. Multiplying the optimal
largest eigenvalue by $d\bar\lambda\eta^2/2$ gives
\eqref{eq:aligned-minimax-value}.

Finally, the condition $\rmM\succeq\alpha^2\vg\vg^\top$ is the second-moment form
of preserving the aligned mean. If $\rmM$ satisfies it, set
$\rmC=\rmM-\alpha^2\vg\vg^\top\succeq0$ and take any zero-mean random vector
$\rvepsilon$ with covariance $\rmC$; then $\rvx=\alpha\vg+\rvepsilon$ has
$\E[\rvx]=\alpha\vg$ and $\E[\rvx\rvx^\top]=\rmM$.

\subsection{Proof of Corollary~\ref{cor:zo-exposure-interpolation}}

Applying Theorem~\ref{thm:anisotropy} with $\rmH$ replaced by
$\vg\vg^\top$ gives
\begin{equation}
  \rmM_q^{\text{ZO}}
  =
  \E\left[\rmP\vg\vg^\top\rmP\right]
  =
  (1-\tau)\vg\vg^\top
  +
  \tau\frac{\norm{\vg}^{2}}{d}\rmI,
\end{equation}
which proves \eqref{eq:zo-exposure-interpolation}. The eigenvalue of
$\rmM_q^{\text{ZO}}$ along $\vg$ is
\begin{equation}
  \norm{\vg}^{2}\left(1-\tau+\frac{\tau}{d}\right),
\end{equation}
and each perpendicular eigenvalue is
$\tau\norm{\vg}^{2}/d$, so the displayed value is the largest eigenvalue. The FO
exposure $\rmM^{\text{FO}}=\vg\vg^\top$ has largest eigenvalue
$\norm{\vg}^{2}$, and $\rmM^\star$ has largest eigenvalue
$\norm{\vg}^{2}/d$. Applying \eqref{eq:hblind-worst-exposure},
\begin{equation}
  \mathcal{R}_{\bar\lambda}\!\left(\rmM_q^{\text{ZO}}\right)
  -
  \mathcal{R}_{\bar\lambda}\!\left(\rmM^\star\right)
  =
  \frac{\eta^2}{2}d\bar\lambda\norm{\vg}^{2}
  (1-\tau)\left(1-\frac{1}{d}\right),
\end{equation}
while
\begin{equation}
  \mathcal{R}_{\bar\lambda}\!\left(\rmM^{\text{FO}}\right)
  -
  \mathcal{R}_{\bar\lambda}\!\left(\rmM^\star\right)
  =
  \frac{\eta^2}{2}d\bar\lambda\norm{\vg}^{2}
  \left(1-\frac{1}{d}\right).
\end{equation}
The first display is therefore $(1-\tau)$ times the second, proving
\eqref{eq:zo-exposure-gap-closing}.
Subtracting
$\E[\rmP\vg]\E[\rmP\vg]^\top$ from
\eqref{eq:zo-exposure-interpolation} also gives the centered covariance used
for covariance-matched controls:
\begin{equation}
  \operatorname{Cov}(\rmP\vg)
  =
  \frac{1}{q+d+1}
  \left(\vg\vg^\top+\norm{\vg}^{2}\rmI\right).
\end{equation}

\subsection{Proof of Theorem~\ref{thm:rise-blockwise-gap}}

For a diagonal block, Theorem~\ref{thm:anisotropy} applies inside the
$d_b$-dimensional block with query count $q_b$:
\begin{equation}
  \E[\rmP_b^\top\rmH_{bb}\rmP_b]
  =
  (1-\tau_b)\rmH_{bb}
  +
  \tau_b\lamavg_b\rmI_b .
\end{equation}
For an off-diagonal block, independence across blocks and
$\E[\rmP_b]=a_b\rmI_b$ give
\begin{equation}
  \E[\rmP_b^\top\rmH_{bc}\rmP_c]
  =
  \E[\rmP_b]^\top\rmH_{bc}\E[\rmP_c]
  =
  a_ba_c\rmH_{bc}.
\end{equation}
This proves \eqref{eq:rise-blockwise-curvature}. Expanding the quadratic form
by blocks,
\begin{equation}
  \mathcal{Q}^{\text{FO}}
  =
  \frac{\eta^2}{2}
  \left[
  \sum_{b=1}^{B}\vg_b^\top\rmH_{bb}\vg_b
  +
  2\sum_{b<c}\vg_b^\top\rmH_{bc}\vg_c
  \right],
\end{equation}
and substituting \eqref{eq:rise-blockwise-curvature} into
$\E[\mathcal{Q}^{\text{\ours}}_{\text{blk}}]$ gives the same expression with
$\rmH_{bb}$ replaced by
$(1-\tau_b)\rmH_{bb}+\tau_b\lamavg_b\rmI_b$ and $\rmH_{bc}$ replaced by
$a_ba_c\rmH_{bc}$. Subtracting yields
\eqref{eq:rise-blockwise-gap}.

\subsection{Proof of Proposition~\ref{prop:rise-blockwise-deviation}}

Let $\rmD_b\triangleq\rmZ_b-\rmI_b$. Applying the same Gaussian
sample-covariance concentration used in
\eqref{eq:app-operator-covariance} to each block and taking a union bound gives
an event of probability at least $1-\sum_b\delta_b$ on which
\begin{equation}
  \norm{\rmD_b}_{\text{op}}
  \le
  C\psi_b(\delta_b)
  \qquad
  \text{for all }b=1,\ldots,B .
  \label{eq:block-sample-cov-event}
\end{equation}
Increasing $C$ absorbs universal constants in the following bounds. On this
event, the diagonal block fluctuation follows from the blockwise version of
\eqref{eq:finite-query-operator-decomp}:
\begin{equation}
  \rmP_b^\top\rmH_{bb}\rmP_b
  -
  \E[\rmP_b^\top\rmH_{bb}\rmP_b]
  =
  \kappa_b^{-1}
  \left[
  \rmD_b\rmH_{bb}
  +
  \rmH_{bb}\rmD_b
  +
  \rmD_b\rmH_{bb}\rmD_b
  -
  \frac{1}{q_b}\left(\rmH_{bb}+\tr(\rmH_{bb})\rmI_b\right)
  \right].
\end{equation}
Here $\rmH_{bb}$ is only assumed symmetric. Thus the centering term is bounded
using
$|\tr(\rmH_{bb})|\le d_b\norm{\rmH_{bb}}_{\text{op}}$, not a PSD trace bound:
\begin{equation}
  \left\|
  \frac{1}{q_b}\left(\rmH_{bb}+\tr(\rmH_{bb})\rmI_b\right)
  \right\|_{\text{op}}
  \le
  \frac{d_b+1}{q_b}\norm{\rmH_{bb}}_{\text{op}}
  \le
  \psi_b(\delta_b)\norm{\rmH_{bb}}_{\text{op}}.
\end{equation}
Together with \eqref{eq:block-sample-cov-event}, this gives
\begin{equation}
  \left\|
  \rmP_b^\top\rmH_{bb}\rmP_b
  -
  \E[\rmP_b^\top\rmH_{bb}\rmP_b]
  \right\|_{\text{op}}
  \le
  C\norm{\rmH_{bb}}_{\text{op}}
  \left(\psi_b(\delta_b)+\psi_b(\delta_b)^2\right).
  \label{eq:block-diagonal-fluctuation}
\end{equation}
For an off-diagonal block, use $\rmP_b=a_b(\rmI_b+\rmD_b)$ to write
\begin{equation}
  \rmP_b^\top\rmH_{bc}\rmP_c-a_ba_c\rmH_{bc}
  =
  a_ba_c
  \left(
  \rmD_b\rmH_{bc}
  +
  \rmH_{bc}\rmD_c
  +
  \rmD_b\rmH_{bc}\rmD_c
  \right).
\end{equation}
Therefore,
\begin{equation}
  \left\|
  \rmP_b^\top\rmH_{bc}\rmP_c-a_ba_c\rmH_{bc}
  \right\|_{\text{op}}
  \le
  C a_ba_c\norm{\rmH_{bc}}_{\text{op}}
  \left(\psi_b(\delta_b)+\psi_c(\delta_c)
  +\psi_b(\delta_b)\psi_c(\delta_c)\right).
  \label{eq:block-cross-fluctuation}
\end{equation}
Finally expand
$\mathcal{Q}^{\text{\ours}}_{\text{blk}}
-\E[\mathcal{Q}^{\text{\ours}}_{\text{blk}}]$ by diagonal and off-diagonal
blocks. Bound each diagonal scalar by
$\norm{\vg_b}^{2}$ times \eqref{eq:block-diagonal-fluctuation}, and each
off-diagonal scalar by $\norm{\vg_b}\norm{\vg_c}$ times
\eqref{eq:block-cross-fluctuation}. The factor $\eta^2/2$ applies to diagonal
terms and the factor $\eta^2$ to $b<c$ cross terms, giving
\eqref{eq:rise-blockwise-deviation}.

\subsection{Proof of Theorem~\ref{thm:rise-blockwise-mean}}

Write $\vDelta=\vm+\veps$, where
$\vm=\E[\vDelta]$ and $\E[\veps]=\vzero$. By the trace identity,
\begin{equation}
  \E\left[\mathcal{Q}(\vDelta)\right]
  =
  \frac{1}{2}\E\left[(\vm+\veps)^\top
  \rmH(\vm+\veps)\right]
  =
  \frac{1}{2}\vm^\top\rmH\vm
  +
  \frac{1}{2}\tr\!\left(
  \rmH\,\E[\veps\veps^\top]
  \right),
\end{equation}
because the cross term has expectation zero. The block independence and
zero cross-covariance assumptions give
\begin{equation}
  \E[\veps\veps^\top]
  =
  \operatorname{blkdiag}(\rmSigma_1,\ldots,\rmSigma_B)
  =
  \rmSigma .
\end{equation}
Using the block trace identity then yields
\begin{equation}
  \tr(\rmH\rmSigma)
  =
  \sum_{b=1}^{B}\tr(\rmH_{bb}\rmSigma_b),
\end{equation}
which proves \eqref{eq:rise-blockwise-mean}.

\subsection{Proof of Proposition~\ref{prop:block-coupling}}

Let
\begin{equation}
  \rmK
  \triangleq
  \rmH_{\text{blk}}^{-1/2}\rmE\rmH_{\text{blk}}^{-1/2}.
\end{equation}
Then
\begin{equation}
  \rmH_{\text{blk}}^{-1/2}\rmH\rmH_{\text{blk}}^{-1/2}
  =
  \rmI+\rmK .
\end{equation}
Since $\norm{\rmK}_{\text{op}}=\epsilon_{\text{blk}}<1$ and $\rmK$ is
symmetric,
\begin{equation}
  (1-\epsilon_{\text{blk}})\rmI
  \preceq
  \rmI+\rmK
  \preceq
  (1+\epsilon_{\text{blk}})\rmI .
\end{equation}
In particular, $\rmI+\rmK\succ0$, and congruence by
$\rmH_{\text{blk}}^{1/2}$ gives $\rmH\succ0$.
Inverting the positive definite matrices reverses the Loewner order:
\begin{equation}
  (1+\epsilon_{\text{blk}})^{-1}\rmI
  \preceq
  (\rmI+\rmK)^{-1}
  \preceq
  (1-\epsilon_{\text{blk}})^{-1}\rmI .
\end{equation}
Finally,
\begin{equation}
  \rmH^{-1}
  =
  \rmH_{\text{blk}}^{-1/2}
  (\rmI+\rmK)^{-1}
  \rmH_{\text{blk}}^{-1/2},
\end{equation}
and congruence by $\rmH_{\text{blk}}^{-1/2}$ gives
\eqref{eq:block-inverse-perturbation}.

\end{document}